\documentclass[lettersize,journal]{IEEEtran}
\usepackage{amsmath,amsfonts}
\usepackage{amssymb}
\usepackage[ruled,vlined,linesnumbered]{algorithm2e}
\usepackage{array}
\usepackage[caption=false,font=footnotesize,labelfont=sf,textfont=sf]{subfig}
\usepackage{booktabs}
\usepackage{multirow}
\usepackage{multicol}
\usepackage{url}
\usepackage{graphicx}
\usepackage{tabularx}
\usepackage{xspace}
\usepackage{cite}
\usepackage{comment}
\usepackage{soul}
\usepackage{hyperref}
\usepackage[dvipsnames]{xcolor}
\usepackage[most]{tcolorbox}
\usepackage{pifont}
\usepackage{listings}
\lstdefinelanguage{systemprompt}{
    basicstyle=\ttfamily\scriptsize,
    breaklines=true,
    frame=none,
    backgroundcolor=\color{white!98!gray},
    breakindent=0pt,
}
\usepackage{colortbl}
\colorlet{Yes}{blue!15}
\colorlet{No}{red!20}
\newcommand{\Name}{\textsc{NeLV}\xspace}
\newcommand{\com}[1]{\tiny$\pm$#1}
\newcolumntype{a}{>{\columncolor{gray!15}}c}

\usepackage[inline]{enumitem}
\newenvironment{tableitemize}
{ \begin{minipage}[t]{\linewidth} \vspace{-5pt} \begin{itemize}[leftmargin=8pt] }
{  \vspace{3pt} \end{itemize} \end{minipage}   }

\def\BibTeX{{\rm B\kern-.05em{\sc i\kern-.025em b}\kern-.08em
    T\kern-.1667em\lower.7ex\hbox{E}\kern-.125emX}}
\makeatletter
\def\subsubsection{\@startsection{subsubsection}{3}{1.25\parindent}{0.1ex plus 0.1ex minus 0.1ex}%
{0.1ex}{\normalfont\normalsize\itshape}}%
\makeatother

\begin{document}

\title{Next-Generation LLM for UAV: \\
From Natural Language to Autonomous Flight}

\author{Liangqi Yuan$^{*}$,~\IEEEmembership{Student Member,~IEEE,}
        Chuhao Deng$^{*}$,~\IEEEmembership{Student Member,~IEEE,}
        Dong-Jun Han,~\IEEEmembership{Member,~IEEE,}
        Inseok Hwang,~\IEEEmembership{Senior Member,~IEEE,}
        Sabine Brunswicker,
        and Christopher G. Brinton,~\IEEEmembership{Senior Member,~IEEE}

\thanks{$^{*}$ Equal contribution}
\thanks{This work was supported in part by the National Science Foundation (NSF) under grants CNS-2146171 and CPS-2313109, the Office of Naval Research (ONR) under grant N00014-22-1-2305, the Air Force Office of Scientific Research (AFOSR) under grant FA9550-24-1-0083, and the Institute of Information \& Communications Technology Planning \& Evaluation (IITP) grant funded by the Korea government (MSIT) (IITP-2026-RS-2026-25544188, Leading Generative AI Human Resources Development).}
\thanks{L. Yuan and C. G. Brinton are with the School of Electrical and Computer Engineering, Purdue University, West Lafayette, IN 47907, USA.  E-mail: liangqiy@purdue.edu; cgb@purdue.edu}
\thanks{C. Deng and I. Hwang are with the School of Aeronautics and Astronautics, Purdue University, West Lafayette, IN 47907, USA.  E-mail: deng113@purdue.edu; ihwang@purdue.edu}
\thanks{D.-J. Han is with the Department of Computer Science and Engineering, Yonsei University, Seoul 03722, South Korea. E-mail: djh@yonsei.ac.kr}
\thanks{S. Brunswicker is with the Polytechnic Institute, Purdue University, West Lafayette, IN 47907, USA. E-mail: sbrunswi@purdue.edu}
}

\maketitle

\begin{abstract}

Large Language Models (LLMs) are increasingly applied to Unmanned Aerial Vehicle (UAV) operations, though mostly to isolated components of small platforms, such as path planning for toy drones, leaving medium- and long-range UAV systems in real-world operational contexts largely uninvestigated. These larger platforms additionally face airport take-off and landing procedures, complex regulatory frameworks, and elevated mission expectations.
This position paper presents the Next-Generation LLM for UAV (\Name) system, which serves as both a working demonstration and an automation roadmap. \Name translates natural language instructions into executable flight trajectories for short-, medium-, and long-range missions through five key technical components: (i) LLM-as-Parser for instruction interpretation, (ii) Route Planner for Points of Interest (POI) determination, (iii) Path Planner for waypoint generation, (iv) Control Platform for trajectory generation, and (v) UAV monitoring.
We demonstrate the system through three representative use cases: multi-UAV patrol, multi-POI delivery, and multi-hop relocation. The system reaches a mission success rate above 97\% on both medium- and long-range missions over 2000 airport pairs across the continental United States, and completes this translation in under three seconds. We further establish a five-level automation taxonomy from LLM-as-Parser (Level 1) to LLM-as-Autopilot (Level 5), identifying the technical prerequisites and research challenges at each stage. Project website: \url{https://liangqiyuan.github.io/NeLV/}.

\end{abstract}


\section{Introduction}

The rise of Large Language Models (LLMs) has transformed numerous domains, such as mobile services, vehicles, transportation, and robotics \cite{mahmud2025integrating, yuan2026device, movahedi2024crossroads, yuan2026large}. These fields have become increasingly intelligent and user-friendly through LLM integration, enabling command and control through natural language. This conversation-based control between humans and LLMs improves both the ability of LLMs to interpret context and the convenience with which humans can direct LLMs to perform actions \cite{yuan2025llmap}. LLMs fulfill diverse roles within these systems. LLM-as-Router can orchestrate task allocation and model selection for human pilots, LLM-as-Agent can execute actions on behalf of humans, and LLM-as-Judge can conduct evaluations in place of human judgment. The increasing specialization of LLMs in distinct roles throughout systems optimizes their performance through clear responsibility allocation. Consequently, LLMs are becoming an essential component of next generation autonomous control and self-driving technologies.

\begin{figure}[t]
    \centering
    \includegraphics[width=1\linewidth]{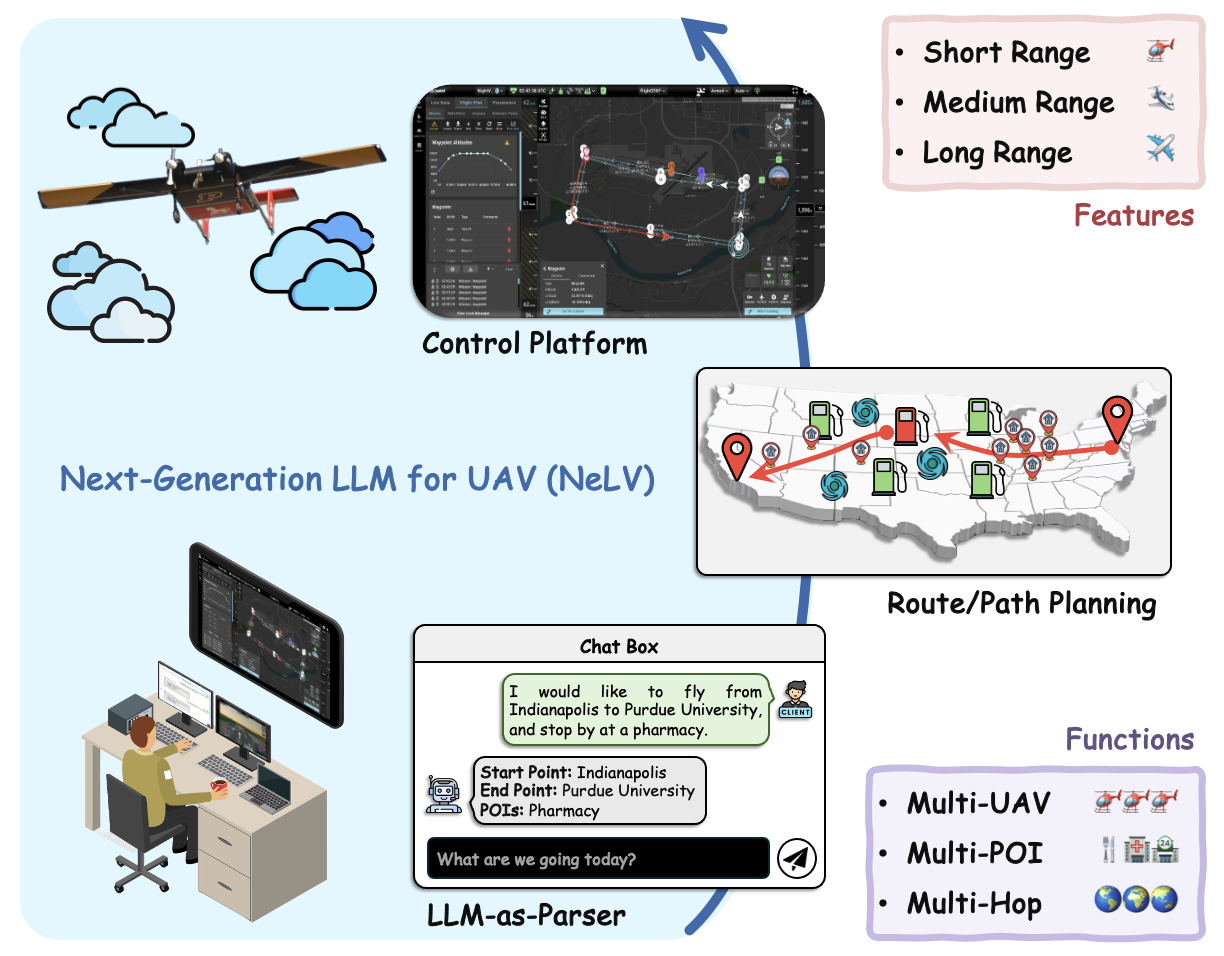}
\vspace{-4mm}
    \caption{Overview of the \Name System.}
\vspace{-4mm}
    \label{fig:Overview}
\end{figure}

\begin{table*}[t]
\renewcommand{\arraystretch}{1.3}
\setlength{\tabcolsep}{4pt}
\centering
\caption{Comparison of Related Literature of LLM for UAV System}
\vspace{-2mm}
\label{Table Comparison of related literature}
\resizebox{\linewidth}{!}{
\begin{tabular}{|l|l|p{1.3cm}|p{3.9cm}|p{3.25cm}|c|c|c|}
\hline
\textbf{Literature} & \textbf{Year} & \textbf{Type} & \textbf{Key Topic} & \textbf{Implementation} & \textbf{Short-Range} & \textbf{Medium-Range} & \textbf{Long-Range} \\
\hline
\hline
\cite{javaid2024large} & 2024 & Survey & LLMs for UAVs & \cellcolor{No} \centering - & \cellcolor{No} - & \cellcolor{No} - & \cellcolor{No} - \\
\hline
\cite{tian2025uavs} & 2025 & Survey & LLMs for UAVs & \cellcolor{No} \centering - & \cellcolor{No} - & \cellcolor{No} - & \cellcolor{No} - \\
\hline

\cite{liu2024large} & 2024 & Survey & LLMs for Transportation & \cellcolor{No} \centering - & \cellcolor{No} - & \cellcolor{No} - & \cellcolor{No} - \\
\hline
\cite{wandelt2024large} & 2024 & Survey & LLMs for Transportation & \cellcolor{No} \centering - & \cellcolor{No} - & \cellcolor{No} - & \cellcolor{No} - \\
\hline

\cite{liu2024generative} & 2024 & Survey & Generative AI for UAVs & \cellcolor{No} \centering - & \cellcolor{No} - & \cellcolor{No} - & \cellcolor{No} - \\
\hline
\cite{sun2024generative} & 2024 & Survey & Generative AI for UAV Networks & \cellcolor{No} \centering - & \cellcolor{No} - & \cellcolor{No} - & \cellcolor{No} - \\
\hline
\cite{kaleem2024emerging} & 2024 & Survey & Generative AI for UAV Networks & \cellcolor{No} \centering - & \cellcolor{No} - & \cellcolor{No} - & \cellcolor{No} - \\
\hline

\cite{yao2024aeroverse} & 2024 & Benchmark & Simulation Platform & \cellcolor{No}Simulator & \cellcolor{Yes}City UAV Simulator & \cellcolor{No} - & \cellcolor{No} - \\
\hline

\cite{liu2023aerialvln} & 2023 & Research & LLM for UAV Planning & \cellcolor{No}Simulator & \cellcolor{Yes}Mission Planning & \cellcolor{No} - & \cellcolor{No} - \\
\hline
\cite{sinha2024real} & 2024 & Research & LLM for UAV Planning & \cellcolor{Yes}Indoor Small UAV & \cellcolor{Yes}Object Detection & \cellcolor{No} - & \cellcolor{No} - \\
\hline
\cite{zhong2024safer} & 2024 & Research & LLM for UAV Planning & \cellcolor{Yes}Indoor Small UAV & \cellcolor{Yes}Obstacle Prediction & \cellcolor{No} - & \cellcolor{No} - \\
\hline
\cite{cui2024tpml} & 2024 & Research & LLM for UAV Planning & \cellcolor{Yes}Indoor Small UAV & \cellcolor{Yes}Mission Planning & \cellcolor{No} - & \cellcolor{No} - \\
\hline
\cite{pueyo2024clipswarm} & 2024 & Research & LLM for UAV Planning & \cellcolor{Yes}Outdoor Small UAV Swarm & \cellcolor{Yes}Formation & \cellcolor{No} - & \cellcolor{No} - \\
\hline
\cite{chen2025typefly} & 2025 & Research & LLM for UAV Planning & \cellcolor{Yes}Indoor Small UAV & \cellcolor{Yes}Mission Planning & \cellcolor{No} - & \cellcolor{No} - \\

\hline
\hline
\cellcolor{Yes}\textbf{Ours} & \cellcolor{Yes}2026 & \cellcolor{Yes}Position Paper 
& \cellcolor{Yes}
\begin{tableitemize}
    \item LLM-as-Parser
    \item Route Planning
    \item Path Planning
    \item Control Platform
    \item UAV Monitoring
\end{tableitemize}
& \cellcolor{Yes}Outdoor Medium UAV
& \cellcolor{Yes}Multi-UAV Patrol & \cellcolor{Yes}Multi-POI Delivery & \cellcolor{Yes}Multi-Hop Relocation \\
\hline
\end{tabular}
}
\vspace{-4mm}
\end{table*}

Unmanned Aerial Vehicles (UAVs) represent an important component of next generation transportation \cite{ahmed2024state, wandelt2024toward, sun2025laerace}. According to data from the Federal Aviation Administration (FAA), registered UAVs in the United States exceed 1 million as of March 2025, with 427,335 remote pilots certified \cite{FAA_Drones}. As UAV numbers increase, this human pilot workforce proves inadequate and represents one of the limitations preventing UAVs from assuming a more substantial role in transportation. How to enable fewer human pilots to operate a single UAV or allow one human pilot to control multiple UAVs remains an ongoing research question. This challenge necessitates assistance from powerful automation systems to help human pilots process information, comprehend data, suggest operations, execute actions, and communicate effectively. LLMs emerge as strong candidates not only because they accept language input, which proves considerably more efficient than alternative command inputs, but also because their robust reasoning capabilities enable them to process UAV-related information effectively.

Although considerable literature focuses on LLM for UAV systems, most studies remain confined to singular aspects, such as using LLMs solely for small toy UAV planning, as detailed in Table \ref{Table Comparison of related literature}. A significant research gap exists regarding a comprehensive system that progresses from human language input to UAV control. These studies do not consider medium- and long-range UAVs, which differ substantially from small platforms. Small UAVs, typically multirotors, can take off from any location and perform short-distance missions, but possess limited cargo capacity and lower performance capabilities. Medium- and long-range UAVs, typically fixed-wings, require airport infrastructure and the adherence to strict take-off and landing procedures. They must also comply with FAA regulations, including prohibitions against flying over crowds and requirements to avoid air traffic control zones. These considerations present new and unique challenges for LLM for UAV systems.

In this paper, we present the Next-Generation LLM for UAV (\Name) system, illustrated in Figure \ref{fig:Overview}. This LLM-powered UAV system translates human language input into executable flight trajectories for short-, medium-, and long-range UAVs missions. Human pilots can access the input for each component and intervene to control or modify operations. While we envision future systems where LLMs play central roles in planning, control, and decision-making, current LLM capabilities remain limited for safety-critical UAV operations. Therefore, this position paper serves dual purposes: (1) \textit{System Demonstration:} We present a working implementation where LLMs handle natural language interpretation (Level 1 of our taxonomy) while traditional algorithms perform route planning, path planning, and control; and (2) \textit{Automation Roadmap:} We establish a five-level taxonomy that defines technical requirements, identifies key challenges, and charts development pathways from current LLM-as-Parser capabilities to future LLM-as-Autopilot systems capable of fully autonomous decision-making.

The contributions of this position paper are as follows.

\begin{itemize}
    \item \textbf{End-to-End System Framework:} We introduce \Name, a framework integrating LLMs with multi-scale UAV operations from natural language input to executable flight trajectories, and to our knowledge the first LLM-powered UAV system capable of supporting short-, medium-, and long-range operations, as well as complex missions involving multi-UAV patrol, multi-POI delivery, and multi-hop relocation scenarios.

    \item \textbf{Five-Stage Pipeline:}  We develop a pipeline in which the LLM interprets the pilot's instruction, classical optimization and control algorithms carry out planning and trajectory generation, and the safety pilot supervises execution: (i) \textit{LLM-as-Parser} that engages in conversational interaction with human pilots to iteratively refine flight plans and interpret operational preferences; (ii) \textit{Route Planner} that considers pilot-specified constraints and multi-objective optimization criteria; (iii) \textit{Path Planner} that circumvents restricted zones, including controlled airspace and adverse meteorological conditions; (iv) \textit{Control Platform} that generates executable trajectories incorporating airport-specific take-off and landing patterns; and (v) \textit{UAV Monitor} that provides real-time mission execution with safety pilot intervention capabilities.

    \item \textbf{Multi-Scale Mission Demonstration and Evaluation:} We demonstrate the system's capabilities through three representative use cases spanning different operational scales: (UC1) short-range multi-UAV patrol, wherein five UAVs conduct surveillance of forested areas within a 5 km radius; (UC2) medium-range multi-POI delivery with two delivery objectives; and (UC3) long-range multi-hop relocation from New York to Los Angeles with intermediate refueling stops. The system integrates airspace information, population density data, and weather forecasts to plan routes and paths and to generate executable trajectories at any public-use airport within the continental United States. We evaluate the accuracy of LLM-as-Parser against extraction baselines, the mission success rate over 2000 airport pairs, and per-component latency.

    \item \textbf{Five-Level Automation Roadmap:}  We establish a systematic taxonomy that situates \Name at Level 1 and defines the evolution toward fully autonomous systems: (L1) \textit{LLM-as-Parser} for instruction interpretation, (L2) \textit{LLM-as-Route-Planner} for strategic route optimization, (L3) \textit{LLM-as-Path-Planner} for tactical path determination and collision avoidance, (L4) \textit{LLM-as-Executor} for integrated control system coordination, and (L5) \textit{LLM-as-Autopilot} for fully autonomous flight operations. For each level, we identify specific technical prerequisites, required knowledge bases, and key research challenges.
\end{itemize}

The remainder of this paper is organized as follows. Section \ref{Sec. Related Work} reviews the literature on LLM for UAV systems. Section \ref{Sec. System} introduces the \Name architecture and its five key technical components. Section \ref{Sec. Experiment} presents the experimental configuration, the datasets, and three representative use cases. Section \ref{Sec. Eval} evaluates parsing accuracy, end-to-end planning success, and per-component latency. Section \ref{Sec. Future Direction} discusses five hierarchical levels of future research directions, before Section \ref{Sec. Conclusion} concludes the paper.

\section{Related Work}
\label{Sec. Related Work}

\subsection{LLM for UAV}

 A significant characteristic of LLMs compared to traditional machine learning methods is their powerful reasoning and generalization capabilities that autonomously complete various tasks. This is particularly important for UAVs, including context understanding, perception, planning, decision-making, and more \cite{kurunathan2023machine, ning2023mobile}, a trend that has been reviewed for intelligent transportation more broadly \cite{wandelt2024large, liu2024large}. Additionally, LLMs enable more natural language-based communication with human pilots, reducing operational burden \cite{guo2024integrating} and potentially allowing non-expert users to operate UAVs in the future. Unlike other automated tasks such as autonomous driving, UAVs are categorized as short-range, medium-range, and long-range vehicles, which can be further classified into fixed-wing, multi-rotor, and other configurations \cite{tian2025uavs}. Each type requires different operational approaches. UAV operation presents greater challenges than that of autonomous vehicles due to various factors. These diverse configurations and structures pose unique challenges in the deployment and generalization of LLMs in the UAV domain. This diversity in configurations, structures, and operational methods also results in a lack of large UAV datasets for LLM training, which is a fundamental issue hindering the deep development of LLMs in this field. Currently, most datasets are limited to bird's-eye view detection of ground objects such as people, vehicles, roads, fires, buildings, and more \cite{han2022comprehensive, yao2024aeroverse, gao2024aerial, chu2024towards, wang2025towards}, representing only a minimal portion of UAV missions. Related datasets for UAV control, planning, and human-pilot language dialogue are scarce \cite{jones2023path, tagliabue2024real}.

Therefore, our proposed \Name system aims to address these research gaps. We present a complete system workflow for LLM for UAV, from language input to actual UAV flight control. Unlike related literature that focuses on complex computer vision tasks (such as identifying ground objects), we emphasize UAV planning, control, and flight. Most importantly, we demonstrate that the \Name system can operate short-range, medium-range, and long-range UAVs.

\subsection{UAV Autonomous System}

Flight plan development in UAV autonomous systems constitutes a prerequisite for mission execution. It encompasses five fundamental phases: take-off, en-route flight, mission execution, return flight, and landing \cite{tang2021automated}. This structure proves particularly essential for fixed-wing UAVs and large-scale UAVs, as deviation from these established protocols potentially precipitates severe safety incidents \cite{sribunma2024mixed}. The existing UAV literature predominantly focuses on small toy UAVs, particularly multirotor configurations, or neglects UAV traffic patterns \cite{roberge2018fast, wang2019completion, bashir2023connectivity}. Furthermore, UAV missions in the literature are predetermined by human pilots without addressing real-world flight planning challenges involving ambiguous objectives and numerous alternatives. For example, for refueling operations, pilots encounter multiple airport options with varying fuel costs, where human pilots select options based on costs, quality, safety, or other factors. Subsequently, after selecting the appropriate nodes, the specific path planning requires a reasonable 3D planning based on multiple objectives \cite{peng2024joint, deng20253d}, presenting a unique challenge that must simultaneously accommodate FAA regulations while determining optimal paths from infinite possible waypoints. Moreover, the trajectory builds upon beyond path planning by incorporating traffic patterns for take-off, landing, and mission execution \cite{federal2011airplane}, requiring an appropriate configuration according to airport specifications. Recall the objectives of our \Name system, all three procedural steps must consider various UAV types, flight ranges, missions, and additional variables, introducing significant implementation challenges.

\begin{figure*}[t]
    \centering
    \includegraphics[width=0.8\linewidth]{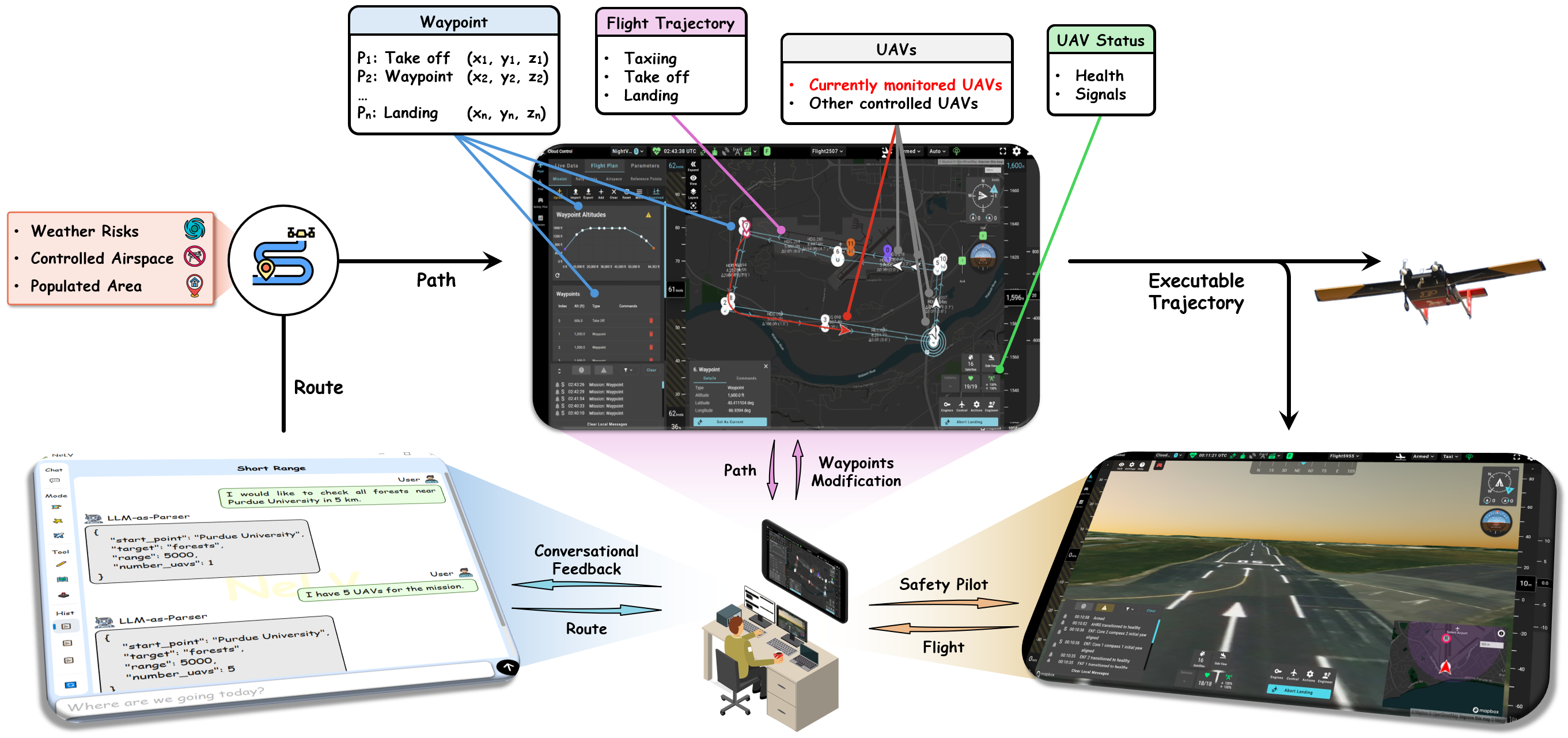}
\vspace{-2mm}
    \caption{Operational Framework and User Interface of the \Name System. \textbf{Left – Chat Box:} The pilot interacts with the LLM through natural-language instructions. After receiving an initial flight plan, the pilot performs route and path planning with visualization support. \textbf{Top – Control Platform \cite{mission_control}:} Once the path is uploaded, the pilot can further refine the mission by manually adjusting waypoints. \textbf{Right – UAV Monitor:} The safety pilot continuously monitors UAVs' status, position, and flight behavior during operation and can make real-time adjustments as needed.} 
\vspace{-2mm}
    \label{Fig. System Framework}
\end{figure*}

The proposed \Name system aims to incorporate and deploy various UAV configurations to accomplish diverse missions. The planning problem progresses hierarchically from high-level node selection to specific path determination, and ultimately to comprehensive trajectory specification. Our \Name system extends the relevant literature by considering how to determine nodes in UAV missions, such as selecting from multiple candidate mission locations. This approach not only enhances mission flexibility, facilitating more adaptable and highly customized flight operations, but also substantially reduces cognitive demands on human pilots through preference-based adaptive flight planning protocols. Furthermore, our \Name system's comprehensive architecture thoroughly addresses all flight components, including take-off, landing, and mission-specific traffic patterns, providing practical solutions for real-world applications.

\let\oldnl\nl
\newcommand{\nlnonumber}{\renewcommand{\nl}{\let\nl\oldnl}}

\begin{algorithm}[t]
\caption{\Name System}
\label{Alg. System}
\small

{\nlnonumber
\begin{center}
    \texttt{/* LLM-as-Parser */}
\end{center}
\vspace{-5pt}}

\textbf{Input:} Pilot instructions \\
\textbf{Output:} Graph $\mathcal{G}$ and constraints $\mathbf{C}^{\max}$ \\
\textbf{Execute:} Algorithm \ref{Alg. LLM} \\
\textbf{Interface:} Chat Box \hfill $\rhd$ Fig. \ref{Fig. System Framework} - Left \\

{\nlnonumber
\begin{center}
    \texttt{/* Route Planner */}
\end{center}
\vspace{-5pt}}

\textbf{Input:} Graph $\mathcal{G}$ and constraints $\mathbf{C}^{\max}$ \\
\textbf{Output:} Route $\xi$ \hfill $\rhd$ Eq. (\ref{Eq. Route}) \\
\textbf{Execute:} Algorithm \ref{Alg. Route} \\
\textbf{Interface:} Chat Box \hfill $\rhd$ Fig. \ref{Fig. System Framework} - Left \\

{\nlnonumber
\begin{center}
    \texttt{/* Path Planner */}
\end{center}
\vspace{-5pt}}

\textbf{Input:} Route $\xi$ \hfill $\rhd$ Eq. (\ref{Eq. Route}) \\
\textbf{Output:} Path $\pi$ \hfill $\rhd$ Eq. (\ref{Eq. Path})  \\
\textbf{Execute:} Algorithm \ref{Alg. Path} \\
\textbf{Interface:} Chat Box \hfill $\rhd$ Fig. \ref{Fig. System Framework} - Top \\

{\nlnonumber
\begin{center}
    \texttt{/* Control Platform */}
\end{center}
\vspace{-5pt}}

\textbf{Input:} Path $\pi$ \hfill $\rhd$ Eq. (\ref{Eq. Path}) \\
\textbf{Output:} Trajectory $\rho$ \hfill $\rhd$ Eq. (\ref{Eq. Trajectory}) \\
\textbf{Execute:} Algorithm \ref{Alg. Circuit} \\
\textbf{Interface:} Control Platform \hfill $\rhd$ Fig. \ref{Fig. System Framework} - Top \\

{\nlnonumber
\begin{center}
    \texttt{/* UAV Monitor */}
\end{center}
\vspace{-5pt}}

\textbf{Input:} Trajectory $\rho$ \hfill $\rhd$ Eq. (\ref{Eq. Trajectory}) \\
\textbf{Execute:} UAV \\
\textbf{Interface:} UAV Monitor \hfill $\rhd$ Fig. \ref{Fig. System Framework} - Right \\

\end{algorithm}

\section{Next-Generation LLM for UAV System}
\label{Sec. System}

\subsection{\Name System Overview}
\label{Sec. System Overview}

As illustrated in Figure \ref{Fig. System Framework}, the \Name system comprises five key components: (i) \textit{LLM-as-Parser} for interpreting human pilot instructions, (ii) \textit{Route Planner} for determining nodes along the flight trajectory, (iii) \textit{Path Planner} for establishing specific paths between each pair of nodes, (iv) \textit{Control Platform} for generating executable trajectories, and (v) \textit{UAV Monitoring} for ultimate flight execution. Pilots retain the capability to intervene and modify system outputs at any stage following component execution through the three interfaces depicted in Figure \ref{Fig. System Framework}, enabling real-time adaptation to suboptimal planning results or evolving mission requirements.

Mathematically, the \Name system operates by constructing a UAV flight graph $\mathcal{G} = (\mathcal{V}, \mathcal{E}, \mathcal{X}, \mathcal{Y})$, where $\mathcal{V}$ represents the set of nodes, $\mathcal{E}$ represents the set of edges, $\mathcal{X}$ is the set of node attributes, and $\mathcal{Y}$ denotes the node types.
\begin{itemize}
    \item[(i)] \textit{LLM-as-Parser} determines nodes $\mathcal{V}$ along with their attributes $\mathcal{X}$ and types $\mathcal{Y}$ based on the UAV mission described by the human pilot, such as forest wildfire surveillance or supermarket supply delivery operations. A specific forest area or supermarket establishment constitutes a node $v \in \mathcal{V}$. These nodes possess distinct attributes $x \in \mathcal{X}$, including geographical coordinates (latitude/longitude), operational status, and congestion levels. The node type $y \in \mathcal{Y}$ represents the designated flight pattern and operational mode, as aircraft may execute aerial cargo drops for supermarket deliveries while airports facilitate conventional landing and take-off procedures.
    \item[(ii)] \textit{Route Planner} performs preliminary route planning based on $\mathcal{G}$, where a route is a sequence of nodes that can be expressed as 
    \begin{equation}
        \xi = [v_1, v_2, \ldots, v_N],
    \label{Eq. Route}
    \end{equation}
    where $v_i \in \mathcal{V}$ and $N \geq 2$ is an integer representing the number of nodes in the route. The route $\xi$ must include take-off and landing nodes, which may be identical. Route planning here is multi-objective, balancing mission completion, flight quality, and cost.
    \item[(iii)] \textit{Path Planner} based on the preliminary route $\xi$ generates a more detailed and specific path: 
    \begin{equation}
    \pi = [v_1, p_{1,2}^{(1)}, p_{1,2}^{(2)}, \ldots, v_2, p_{2,3}^{(1)}, p_{2,3}^{(2)}, \ldots, v_N],
    \label{Eq. Path}
    \end{equation}
    where $p_{i,i+1}^{(j)}$ represents the $j$-th waypoint in the path segment connecting node $v_i$ to node $v_{i+1}$. This detailed path planning avoids the published airspace volumes of each class along the route.
    \item[(iv)] \textit{Control Platform} transforms the path $\pi$ into an executable trajectory:
    \begin{equation}
    \rho = [\mathbf{V}_1, p_{1,2}^{(1)}, p_{1,2}^{(2)}, \ldots, \mathbf{V}_2, p_{2,3}^{(1)}, p_{2,3}^{(2)}, \ldots, \mathbf{V}_N],
    \label{Eq. Trajectory}
    \end{equation}
    where each $\mathbf{V}_i$ represents a specialized air traffic pattern corresponding to the node type $y_i \in \mathcal{Y}$ of node $v_i$. These patterns include specific procedures such as take-off sequences, loiter points, and landing procedures tailored to the operational context of each node.
    \item[(v)] \textit{UAV Execution and Monitoring:} The UAV receives trajectory coordinates and associated control commands from the control platform. The onboard flight control systems autonomously navigate the aircraft to sequential waypoints while maintaining prescribed flight parameters. Concurrently, the UAV's integrated sensor suite transmits real-time telemetry data, including precise geolocation coordinates, fuel levels, engine RPM, oil temperature, and other critical operational parameters, to the ground control platform. Remote operators continuously monitor all system parameters and maintain supervisory control authority to implement corrective interventions when operational anomalies or safety considerations necessitate manual override.
\end{itemize}
Next, we introduce the algorithms of each key component.

\begin{algorithm}[t]
\caption{LLM-as-Parser and Graph Construction}
\label{Alg. LLM}
\small
\nlnonumber \hspace{-10pt} \textbf{Input from Human Pilot:} Initial instruction ($I_0$) and subsequent instructions ($I_t$ for $t > 0$)

\nlnonumber \hspace{-10pt} \textbf{Input from Map Service:} Node attributes ($\mathcal{X}$)

\nlnonumber \hspace{-10pt} \textbf{Output:} Graph ($\mathcal{G}$) and constraints $\mathbf{C}^{\max}$

Initialize sets of nodes $\mathcal{V}$, types $\mathcal{Y}$, and constraints $\mathbf{C}^{\max}$ \\
Initialize conversation $\mathcal{I} \leftarrow I_0$ \\
Initialize time index $t \leftarrow 0$ \\

\While{instruction $I_t$ exists}{
    Concatenate instruction to history $\mathcal{I} \leftarrow \mathcal{I} \oplus I_t$ \\
    LLM reason using conversation $R_t \leftarrow \text{LLM}(\mathcal{I})$ \\
    Extract information from response: $\mathcal{Y}, \mathbf{C}^{\max} \leftarrow R_t$ \\
    Search from Map Service to obtain nodes $\mathcal{V}$ and their attributes $\mathcal{X}$ based on node types $\mathcal{Y}$ \\
    Concatenate response to history $\mathcal{I} \leftarrow \mathcal{I} \oplus R_t$ \\
    $t \leftarrow t + 1$ \\
}

Initialize set of edges $\mathcal{E}$ \\
Obtain number of nodes $N \leftarrow |\mathcal{V}|$ \\

\For{$i = 1, \dots, N$}{
    \For{$j = 1, \dots, N$}{
        \If{$i \neq j$}{
            Compute edge weight $w_{i, j}$ between nodes $v_i$ and $v_j$ \\
            Add edge $\mathcal{E} \leftarrow \mathcal{E} \cup \{(v_i, v_j, w_{i, j})\}$ \\
        }
    }
}

Construct graph $\mathcal{G} \leftarrow (\mathcal{V}, \mathcal{E}, \mathcal{X}, \mathcal{Y})$
\end{algorithm}

\begin{algorithm}[t]
\caption{Route Planner}
\label{Alg. Route}
\small
\nlnonumber \hspace{-10pt} \textbf{Input from LLM-as-Parser:} Graph ($\mathcal{G}$) and constraints ($\mathbf{C}^{\max}$)

\nlnonumber \hspace{-10pt} \textbf{Output:} Optimal route ($\xi^\ast$) 

Initialize candidate route set $\Omega$ \\
\For{each set $\mathbf{y} \subseteq \mathcal{Y}$}{
    \For{permutation $\vec{\mathbf{y}}$ of $\mathbf{y}$}{
        \If{$\vec{\mathbf{y}}$ satisfies constraints $\mathbf{C}^{\max}$}{
            Construct a subgraph $\mathcal{G}_{\vec{\mathbf{y}}}$ containing only nodes of types in $\vec{\mathbf{y}}$ \\
            Search subgraph $\mathcal{G}_{\vec{\mathbf{y}}}$ to find locally optimal route $\widetilde{\xi}$ \hfill $\rhd$ Eq. (\ref{Eq. Route Planning}) \\
            Add $\widetilde{\xi}$ to candidate set: $\Omega \leftarrow \Omega \cup \{\widetilde{\xi}\}$
        }
    }
    \If{$\Omega \neq \emptyset$}{
        Select optimal route $\xi^\ast$ from $\Omega$ based on human pilot or objective function $\mathcal{O}_{\text{route}}(\xi)$ \hfill $\rhd$ Eq. (\ref{Eq. Route Planning}) \\
        Early stop
    }
}
\Else{Output direct route if no feasible solution $\xi^\ast \leftarrow [v_1, v_N]$}
\end{algorithm}

\subsection{LLM-as-Parser}

LLM-as-Parser constitutes the most fundamental component of the \Name system, as it defines the UAV flight graph $\mathcal{G}$, and its accuracy directly impacts the outcomes of all subsequent processes. Although we envision future LLMs possessing capabilities beyond merely defining graph $\mathcal{G}$ to encompass route determination $\xi$, path planning $\pi$, and executable trajectory generation $\rho$, we have determined that current general-purpose LLMs lack sufficient capability due to the absence of specialized LLMs trained on UAV planning datasets. Therefore, we presently utilize LLMs exclusively as instruction parsers to extract essential information from natural language instructions, including departure points, destination coordinates, and mission specifications.

\begin{figure}[t]
\begin{tcolorbox}[colback=white!98!gray, colframe=black!70, enhanced, boxrule=0.5pt, arc=1pt, left=6pt, right=6pt, top=6pt, bottom=6pt, fonttitle=\bfseries\small, title=System Prompt: Short-Range Multi-UAV Patrol]
\vspace{-4mm}
\begin{lstlisting}[language=systemprompt]
Extract POIs, constraints, and weights from human instruction through step-by-step reasoning:

1) **Understand the conversation flow**
   - Determine if this is a new route planning request or a modification to an existing route.
   - For modifications, identify only the specific changes requested and preserve existing parameters.

2) **Identify the key location**
   - Identify and extract the starting point from the user's instruction.

3) **Define target and operational range**
   - Extract the target name and operational range (in meters) from the user's instruction.

4) **Extract UAV count**
    - Count the total number of UAVs mentioned in the instruction.
    - If no specific count is provided, maintain previous value or default to 1.

5) **Ensure consistency and correctness**
   - Always return a valid JSON object without any additional text, explanation, or error messages.
   - Maintain all previously provided values unless explicitly changed by the user.

Output must be valid JSON with structure:
{
    "start_point": string,
    "target": string,
    "range": integer,
    "number_uavs": integer,
}
\end{lstlisting}
\vspace{-4mm}
\end{tcolorbox}
\vspace{-4mm}
\caption{System prompt of LLM-as-Parser for short-range missions. The medium- and long-range prompts open with the same two steps and close with the same consistency check, and the steps between them extract the fields each mission type carries.}
\vspace{-4mm}
\label{Fig. Prompt}
\end{figure}

In a typical operational scenario, a human pilot might instruct: ``I would like to check all forests near Purdue University within 5 km." The LLM-as-Parser systematically extracts ``Purdue University" as the reference point, ``forests" as the surveillance target, and ``5000" as the operational radius. The extracted parameters are returned as a single JSON object whose schema is fixed by the mission type, which for a short-range patrol is \texttt{\{start\_point, target, range, number\_uavs\}}; Fig.~\ref{Fig. Use_Case_1_LLM} shows the parser output for this instruction, in which the fleet size takes its default value because the pilot did not specify one. The system prompt that elicits this object is given in Fig.~\ref{Fig. Prompt}. This parsing process becomes challenging when accounting for inherent variability in human linguistic behavior, including potential typographical errors, linguistic ambiguities, and incomplete information specifications. Human pilots therefore supplement or correct previous instructions through iterative dialogue. Continuing the aforementioned scenario, if the pilot initially omits the number of UAVs required for mission execution, they can subsequently provide clarification by stating: ``I have 5 UAVs for the mission." The parser then returns the same object with \texttt{number\_uavs} updated to 5 and every other field carried over unchanged, since the system prompt instructs the model to preserve previously established values unless the pilot explicitly revises them. These modifications can occur at any stage of the \Name pipeline, from the initial instruction to adjustments made after route or path planning.

\subsection{Route Planning} \label{Sec. Route}

Following the construction of the graph $\mathcal{G}$ by LLM-as-Parser, the next step involves identifying a route within $\mathcal{G}$ from the start point $v_1$ to the end point $v_N$ that fulfills the mission requirements. These key points are represented as nodes $v \in \mathcal{V}$ in $\mathcal{G}$. The fundamental goal of route planning is to determine a sequence of nodes as the route $\xi = [v_1, v_2, \ldots, v_N]$ based on multiple objectives, such as identifying optimal fueling locations according to fuel costs and consumption profiles, while accommodating various constraints, such as time limitations. Mathematically, this optimization problem can be formulated as:
\begin{equation}
\small
\begin{aligned}
&\xi^\ast = \underset{\xi \in \Omega}{\arg\min} \; \mathcal{O}_{\text{route}}(\xi) \\
&\quad \mathcal{O}_{\text{route}}(\xi) = \sum_{i=1}^{N-1} \Big(\alpha \cdot \mathcal{L}(v_i, v_{i+1}) - (1 - \alpha) \cdot \mathcal{R}(v_{i+1}) \Big) \\
&\quad \mathrm{s.t.} \;
\mathbf{C}(\xi) \leq \mathbf{C}^{\max}, \quad \quad \; \; \;M(\xi) = 1 \\
\end{aligned},
\label{Eq. Route Planning}
\end{equation}
where $\xi^\ast$ represents the optimal route, $\Omega$ denotes the feasible route space from start point to end point, and $\mathcal{O}_{\text{route}}(\xi)$ characterizes the objective function governing the route optimization process. $\mathcal{L}(v_i, v_{i+1})$ represents the cost function between consecutive nodes, incorporating factors such as air traffic density, prevailing wind conditions, fuel consumption, etc. $\mathcal{R}(v_{i+1})$ denotes the reward or utility associated with visiting node $v_{i+1}$. The parameter $\alpha$ balances the trade-off between minimizing costs and maximizing rewards. $\mathbf{C}(\xi)$ represents a vector of constraint functions with upper bound $\mathbf{C}^{\max}$, encompassing various operational limitations. $M(\xi)$ signifies the mission completion indicator function, which is equal to $1$ if all mission requirements are satisfied and $0$ otherwise.

These functions are determined by UAV specifications, mission parameters, and operational contexts, and are configured on a case-by-case basis according to specific requirements. For instance, when executing a long-range multi-hop relocation mission requiring intermediate refueling, the cost function $\mathcal{L}(v_i, v_{i+1})$ incorporates fuel pricing variations across different airports and flight-related consumption metrics, while the reward function $\mathcal{R}(v_{i+1})$ may assume negligible values when fuel quality considerations are not operationally relevant. The constraint function $\mathbf{C}(\xi)$ represents current fuel capacity levels and range limitations, while the mission completion criterion $M(\xi)$ indicates successful refueling operations and mission objective fulfillment.

\begin{algorithm}[t]
\caption{Path Planner}
\label{Alg. Path}
\small
\nlnonumber \hspace{-10pt} \textbf{Input from Route Planner:} Route ($\xi$)

\nlnonumber \hspace{-10pt} \textbf{Input from Services:} Service data ($\mathcal{S}$)

\nlnonumber \hspace{-10pt} \textbf{Output:} Optimal path ($\pi^\ast$) 

\For{each edge $(v_i, v_{i+1}) \in \xi$ \textbf{in parallel}}{
Initialize population $\mathcal{P}$, personal best positions $\mathbf{p}_0$, global best $\mathbf{g}_0$, and velocity $\mathbf{v}_0$

\For{$t = 1, \dots, T-1$}{
    Generate random vectors $\mathbf{r}_p$ and $\mathbf{r}_g$ \\
    Update velocity and positions \hfill $\rhd$ Eq. (\ref{Eq. PSO})\\
    Evaluate objective function $\mathcal{O}_{\text{path}}(\mathcal{P}_{t+1})$ \hfill $\rhd$ Eq. (\ref{Eq. Path Planning Instantiated})\\
    \For{$z = 1, \dots, N$}{
        \If{$\mathcal{O}_{\text{path}}(\mathcal{P}_{t+1}[z]) < \mathcal{O}_{\text{path}}(\mathbf{p}_t[z])$}{
            Update personal best: $\mathbf{p}_{t+1}[z] \gets \mathcal{P}_{t+1}[z]$
            }
        \Else{
            Maintain personal best: $\mathbf{p}_{t+1}[z] \gets \mathbf{p}_t[z]$
            }
    }
    Update global best: $\mathbf{g}_{t+1} \gets \arg\min_{p \in \mathbf{p}_{t+1}} \mathcal{O}_{\text{path}}(p)$\\
}
Edge path $\pi_i^\ast \gets \mathbf{g}_T$}
Output optimal path $\pi^\ast \gets$ concatenation of $\pi_1^\ast, \dots, \pi_{|\xi|-1}^\ast$
\end{algorithm}

\subsection{Path Planning} \label{Sec. Path}

Following the determination of the optimal route $\xi^\ast$ by the route planner, the path planner generates a more detailed and specific path $\pi$, accounting for a comprehensive set of service data $\mathcal{S}$, which encompasses weather forecasts, ground risks, and airspace restrictions. Given the nodes from $\xi$, the path planning problem involves finding intermediate waypoints $p^{(j)}_{i,i+1} \in \mathbb{R}^3$ (representing longitude, latitude, and altitude) to construct a path $\pi$ that optimizes a composite objective function. For each segment connecting nodes $v_i$ and $v_{i+1}$, the path planning problem can be formulated as:
\begin{equation}
\pi^\ast_{i,i+1} = \underset{\pi \in \Pi_{i,i+1}}{\arg\min} \; \mathcal{O}_{\text{path}}(\pi),
\label{Eq. Path Planning}
\end{equation}
where $\pi^\ast_{i,i+1}$ designates the optimal path segment between nodes $v_i$ and $v_{i+1}$, $\Pi_{i,i+1}$ is the set of all feasible paths between these nodes, and $\mathcal{O}_{\text{path}}(\pi)$ is the objective evaluated on this path segment with respect to the service data $\mathcal{S}$.  Path planning admits infinitely many solutions over a continuous space of waypoints, so the constraints enter the objective as penalties weighted an order of magnitude above the cost terms, which yields
\begin{equation}
\begin{aligned}
\mathcal{O}_{\text{path}}(\pi) = \; & w_0 \textstyle\sum_{k} \Phi_k(\pi)
+ w_1 \Big( 1 - \frac{L_{\min}}{L(\pi)} \Big) \\
& + w_2 \frac{R_{\text{wx}}(\pi)}{R_{\text{wx}}^{\max}}
+ w_3 \frac{R_{\text{gnd}}(\pi)}{R_{\text{gnd}}^{\max}},
\end{aligned}
\label{Eq. Path Planning Instantiated}
\end{equation}
where $\Phi_k(\pi)$ counts the violations of the turn-angle, airspace, and search-bound constraints, the index $k$ running over these three types, $L(\pi)$ denotes the length of the segment and $L_{\min}$ its lower bound, so that the first cost term is a relative detour ratio in $[0,1)$, while $R_{\text{wx}}(\pi)$ and $R_{\text{gnd}}(\pi)$ are the accumulated weather and ground risks divided by the reference levels $R_{\text{wx}}^{\max}$ and $R_{\text{gnd}}^{\max}$. Normalization renders the three cost terms numerically commensurate but not equally important, and $w_1$, $w_2$, and $w_3$ encode the human pilot's preferred trade-off between mission efficiency and risk exposure.

\begin{algorithm}[t]
\caption{Take-off and Landing Pattern Generator}
\label{Alg. Circuit}
\small
\nlnonumber \hspace{-10pt} \textbf{Input:} latitude-longitude coordinates $(\varphi_s,\lambda_s)$, start altitude ($h_s$), runway heading ($\theta_0$), waypoint separation ($d$), altitude profile ($h$), traffic pattern ($\mathrm{TP}$)

\nlnonumber \hspace{-10pt} \textbf{Output:} Local Circuit ($\mathbf{V}$)

Traffic pattern sign $s \!\gets\! \begin{cases}-1,&\mathrm{TP}=\text{Left}\\+1,&\mathrm{TP}=\text{Right}\end{cases}$\;
Fixed angular offsets $\Theta_{\mathrm{RA}}\!\gets\!s\cdot 90^{\circ}$, $\Theta_{\mathrm{IA}}\!\gets\!s\cdot 12^{\circ}$

\BlankLine
Heading-increment $\boldsymbol{\Delta\theta} \!\gets\!
  \bigl[0,\,0,\,\Theta_{\mathrm{RA}},\,\Theta_{\mathrm{RA}},\,
        \Theta_{\mathrm{RA}},\,\Theta_{\mathrm{IA}},\,
        -2\Theta_{\mathrm{IA}},\,\Theta_{\mathrm{RA}},\,
        \Theta_{\mathrm{RA}},\,\Theta_{\mathrm{RA}},\,0\bigr]$\\

\BlankLine
$\mathbf{V}\gets\bigl\{(\varphi_s,\lambda_s,h_0)\bigr\}$

\BlankLine
\For{$k = 1,\dots,9$}{
    $\theta_k \gets \theta_{k-1} + \Delta\theta_k$\\
    $(\varphi_k,\lambda_k) \gets 
        \mathcal{P}\!\bigl(\varphi_{k-1},\lambda_{k-1};
                           d_k,\theta_k\bigr)$ \hfill $\rhd$ Eq.~(\ref{eqn:circuit})\\
    $\mathbf{V} \gets \mathbf{V} \cup \bigl\{(\varphi_k,\lambda_k,h_k)\bigr\}$
}
\Return $\mathbf{V}$
\end{algorithm}

The complete optimal path $\pi^\ast$ is constructed by concatenating the optimal path segments $\pi^\ast_{i,i+1}$ between consecutive nodes in route $\xi$, yielding $\pi^\ast = v^\ast_1 \oplus \pi^\ast_{1,2} \oplus v^\ast_2 \oplus \pi^\ast_{2,3} \oplus \cdots \oplus \pi^\ast_{N-1,N} \oplus v^\ast_N$, where $\oplus$ denotes path concatenation. To effectively and efficiently determine $\pi$ within a large search space comprising both discrete and continuous variables, we employ Particle Swarm Optimization (PSO) \cite{kennedy1995particle}, a metaheuristic algorithm well suited for this class of problems. For path planning applications, PSO begins by randomly generating a set of candidate paths, collectively forming an initial population $\mathcal{P}_0$. This population evolves iteratively for $T$ generations according to the following update equations:
\begin{equation}
\begin{aligned}
\mathbf{v}_{t+1} &= w \mathbf{v}_{t} + c_1 \mathbf{r}_p \cdot (\mathbf{p}_t - \mathcal{P}_t) + c_2 \mathbf{r}_g \cdot (\mathbf{g}_t - \mathcal{P}_t) \\
\mathcal{P}_{t+1} &= \mathcal{P}_{t} + \mathbf{v}_{t+1}
\end{aligned},
\label{Eq. PSO}
\end{equation}
where $t = 1, 2, \ldots, T$ denotes the generation index. The parameters $w$, $c_1$, and $c_2$ represent inertia weight, personal influence, and social influence, respectively, which are hyperparameters for tuning PSO. The term $\mathbf{p}_t$ represents the personal best positions historically achieved by each particle and $\mathbf{g}_t$ denotes the global best solution identified across all particles at generation $t$. The vectors $\mathbf{r}_p$ and $\mathbf{r}_g$ contain uniformly distributed random numbers in the interval $[0,1]$. The detailed algorithmic implementation of this PSO-based path planner is presented in Algorithm \ref{Alg. Path}.

\begin{figure}[t]
    \centering
    \includegraphics[width=\linewidth]{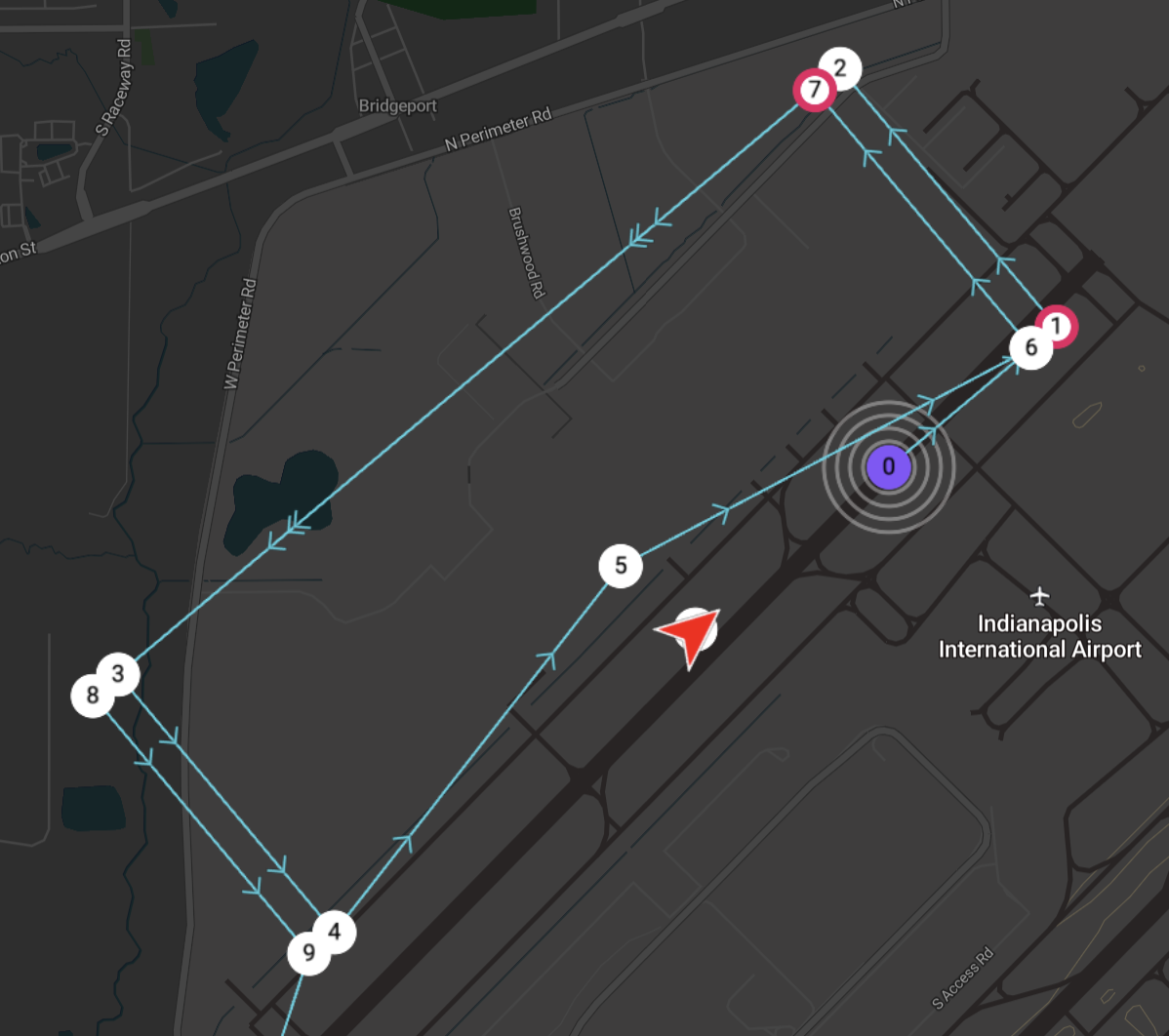}
\vspace{-4mm}
    \caption{Standard take-off circuit pattern configured in a kidney-shaped trajectory to enable the UAV to maintain controlled loitering within the traffic pattern until air traffic control (ATC) clearance is received. The purple waypoint (0) indicates the initial take-off position and white waypoints represent intermediate navigation points along the flight path. The blue connecting lines illustrate the planned trajectory segments between consecutive waypoints, with the directional arrows indicating the prescribed flight direction and sequence. Waypoint 5 is strategically positioned to prevent the aircraft from overflying the active runway, ensuring compliance with aviation safety protocols and airport operational procedures.} 
\vspace{-4mm}
    \label{Fig. Circuit}
\end{figure}

\subsection{Control Platform} 
\label{Sec. Control Platform}

To transform the optimal path $\pi^*$ into a realistic mission profile suitable for UAV deployment, the initial waypoint $v_1$ and terminal waypoint $v_N$ must be replaced with standardized take-off and landing circuit patterns, respectively. Based on the coordinates of $v_1$ and $v_N$, the corresponding airports are identified through geospatial matching algorithms. Subsequently, the coordinates and headings of all runway endpoints are retrieved from the OurAirports database \cite{ourairports} and utilized to compute the runway centerpoint $(\varphi_s,\lambda_s)$, which serves as the reference origin for circuit generation. Given predefined altitude specifications and waypoint separation parameters, a local circuit pattern is systematically generated using Algorithm~\ref{Alg. Circuit}, where the waypoint sequence is indexed by $k=0,1,\dots,8$. The geodetic calculations for circuit waypoint generation begin with the angular distance normalization, where $d_k$ represents the great-circle distance between consecutive waypoints $k$ and $k+1$ in meters:
\begin{equation}
\delta_k = \frac{d_k}{6.371\times10^6}
\label{eqn:angular_distance}
\end{equation}
where $\delta_k$ represents the angular distance normalized by Earth's radius ($6.371\times10^6$ meters). Using this normalized angular distance $\delta$ and the bearing angle $\theta_k$ from waypoint $k-1$ to waypoint $k$, the latitude and longitude coordinates of each waypoint are computed through spherical trigonometry:
\begin{equation} 
\begin{aligned}
    \varphi_k &= \arcsin(\sin\varphi_{k-1}\cos\delta_k + \cos\varphi_{k-1}\sin\delta_k\cos\theta_k) \\
    \lambda_k &= \lambda_{k-1} + \operatorname{atan2}(\sin\theta_k\sin\delta_k\cos\varphi_{k-1}, \\
    &\quad\quad\quad\quad\quad\quad\quad\quad\quad \cos\delta_k - \sin\varphi_{k-1}\sin\varphi_k)
\end{aligned}
\label{eqn:circuit}
\end{equation}
where $\varphi_k$ denotes the latitude and $\lambda_k$ denotes the longitude of the $k$-th waypoint. An exemplary circuit configuration made with \cite{mission_control} demonstrating these geodetic transformations is illustrated in Figure~\ref{Fig. Circuit}. Finally, mission-specific operational commands, such as clockwise loitering maneuvers around designated waypoints for specified durations to facilitate cargo drop-off or loading operations, are systematically appended to construct the final executable trajectory $\rho$, which is subsequently uploaded to the control platform for autonomous mission execution.

\section{Implementation and Case Study}
\label{Sec. Experiment}

\subsection{Graphical User Interface and Setup}

\begin{figure*}[t]
\centering
\begin{minipage}[b]{0.5\textwidth}
    \centering
    \subfloat[LLM-as-Parser\label{Fig. Use_Case_1_LLM}]{
        \includegraphics[height=\linewidth]{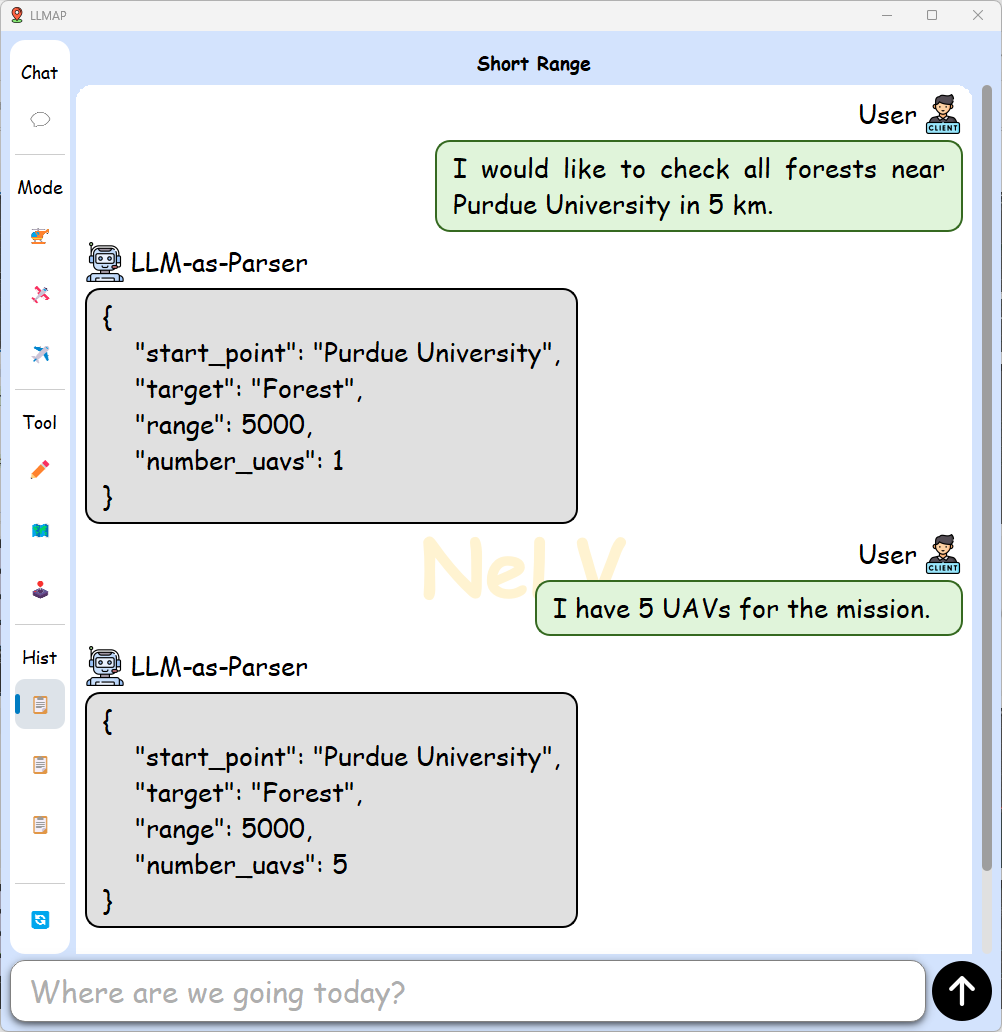}
    }
\end{minipage}
\hfill
\begin{minipage}[b]{0.46\textwidth}
    \subfloat[Route Planning\label{Fig. Use_Case_1_Route}]{
        \includegraphics[width=0.45\linewidth]{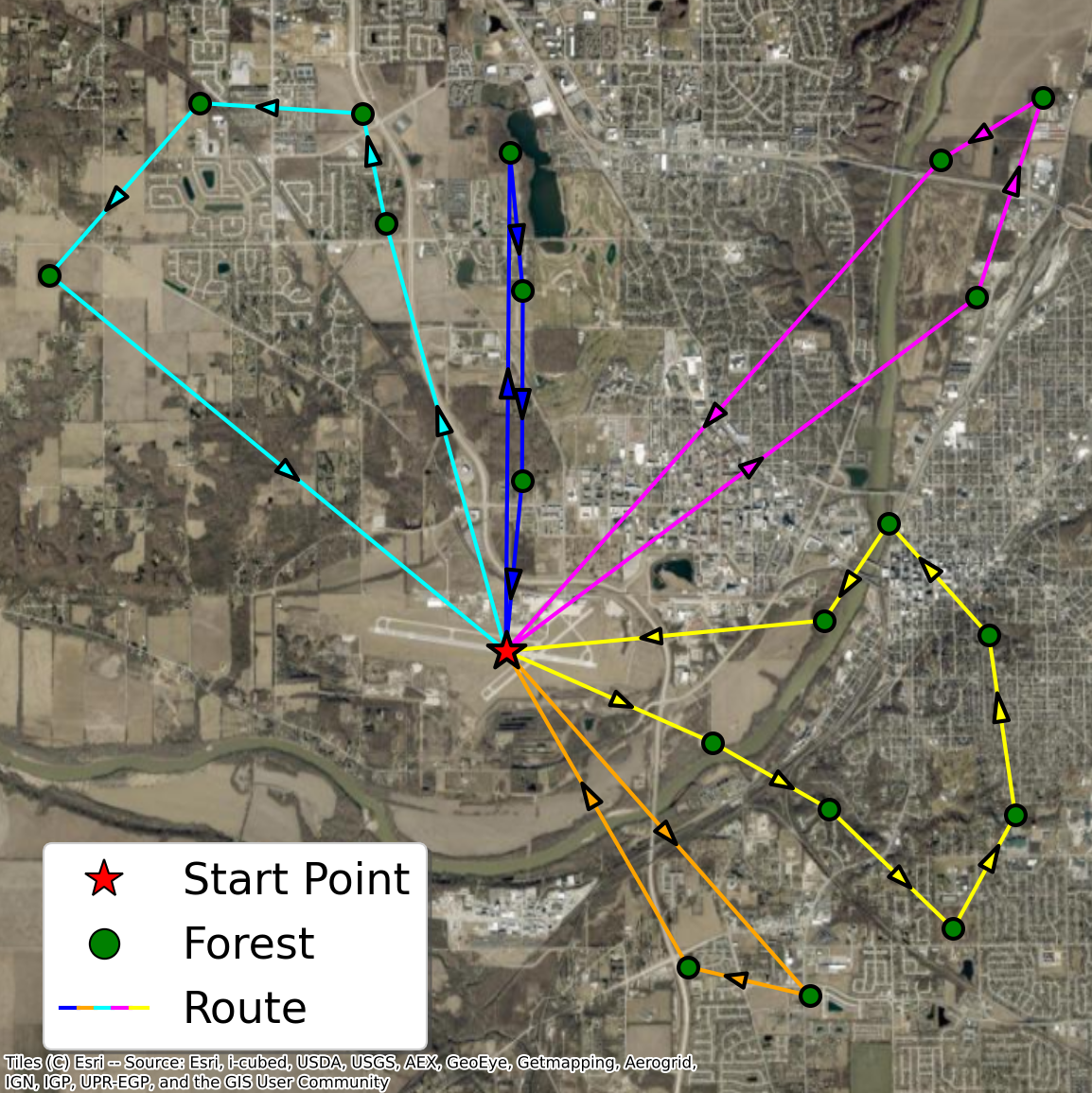}
    }
    \hfill
    \subfloat[Path Planning\label{Fig. Use_Case_1_Path}]{
        \includegraphics[width=0.45\linewidth]{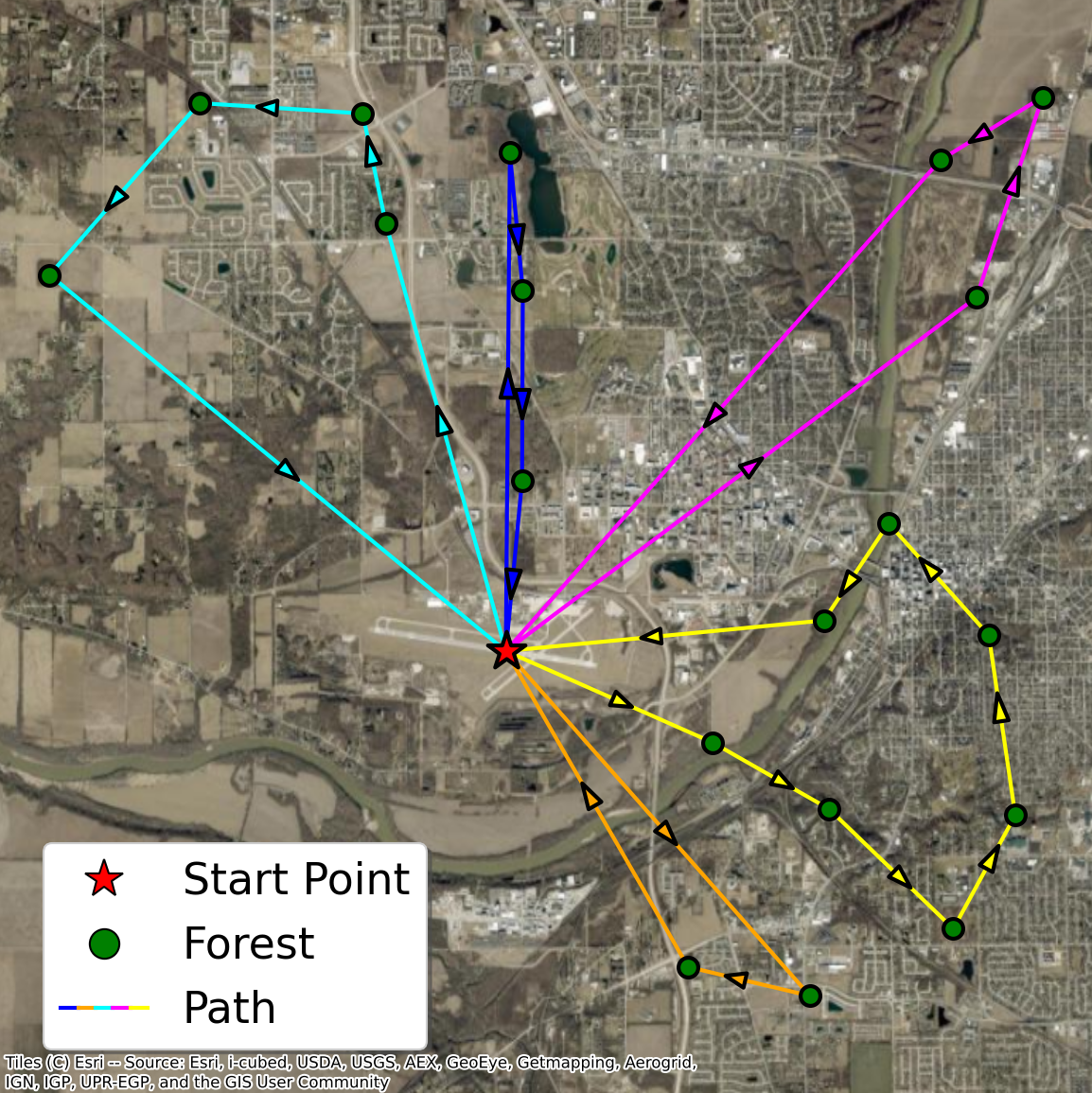}
    }
    \vskip1ex
    \subfloat[Executable Trajectory \label{Fig. Use_Case_1_Traj}]{
        \includegraphics[width=1.0\linewidth]{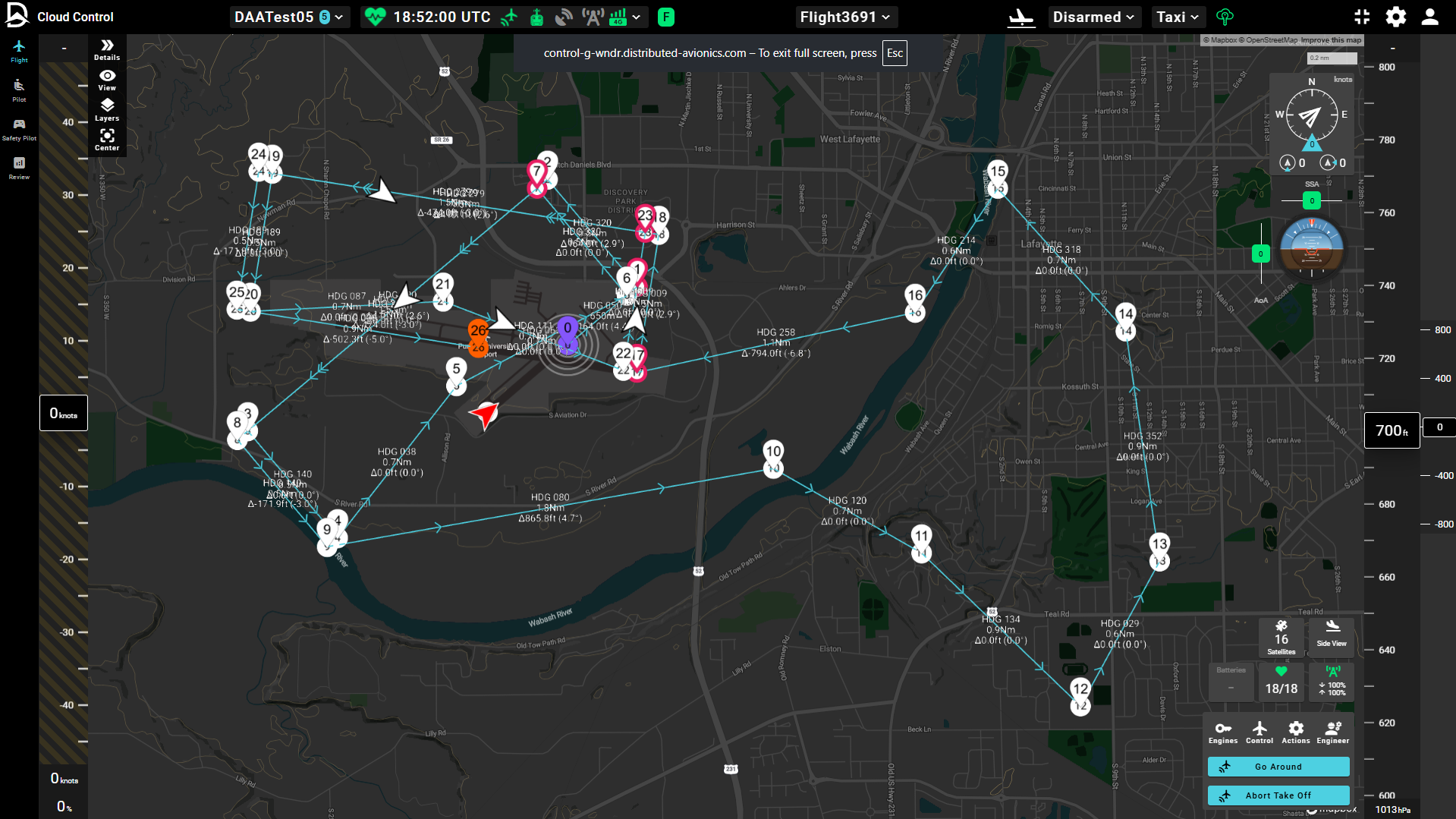}
    }
\end{minipage}
\caption{Use Case 1: Short-Range Multi-UAV Patrol (Sec.~\ref{Sec. Use Case 1}).}
\vspace{-4mm}
\label{Fig. Use_Case_1}
\end{figure*}

\begin{figure*}[t]
\centering
\begin{minipage}[b]{0.50\textwidth}
    \centering
    \subfloat[LLM-as-Parser\label{Fig. Use_Case_2_LLM}]{
        \includegraphics[height=\linewidth]{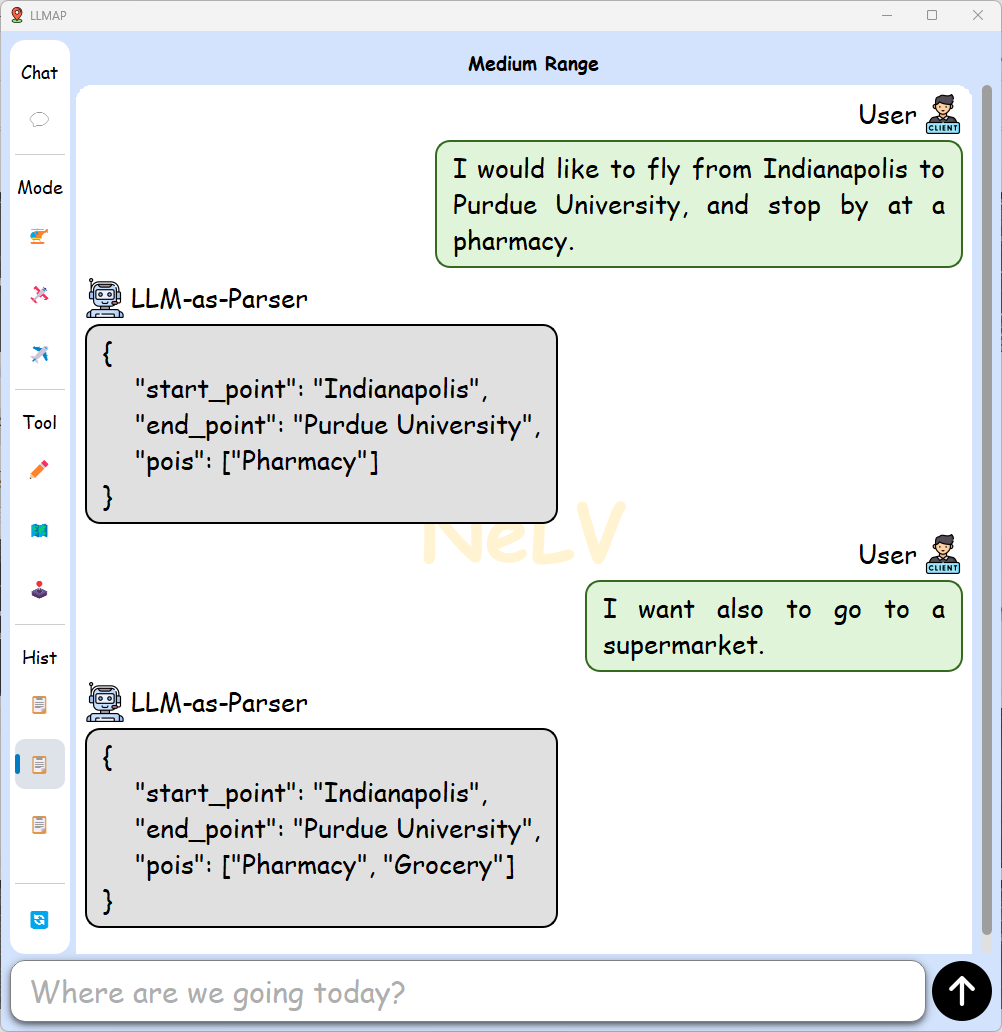}
    }
\end{minipage}
\hfill
\begin{minipage}[b]{0.48\textwidth}
    \subfloat[Route Planning\label{Fig. Use_Case_2_Route}]{
        \includegraphics[width=0.45\linewidth]{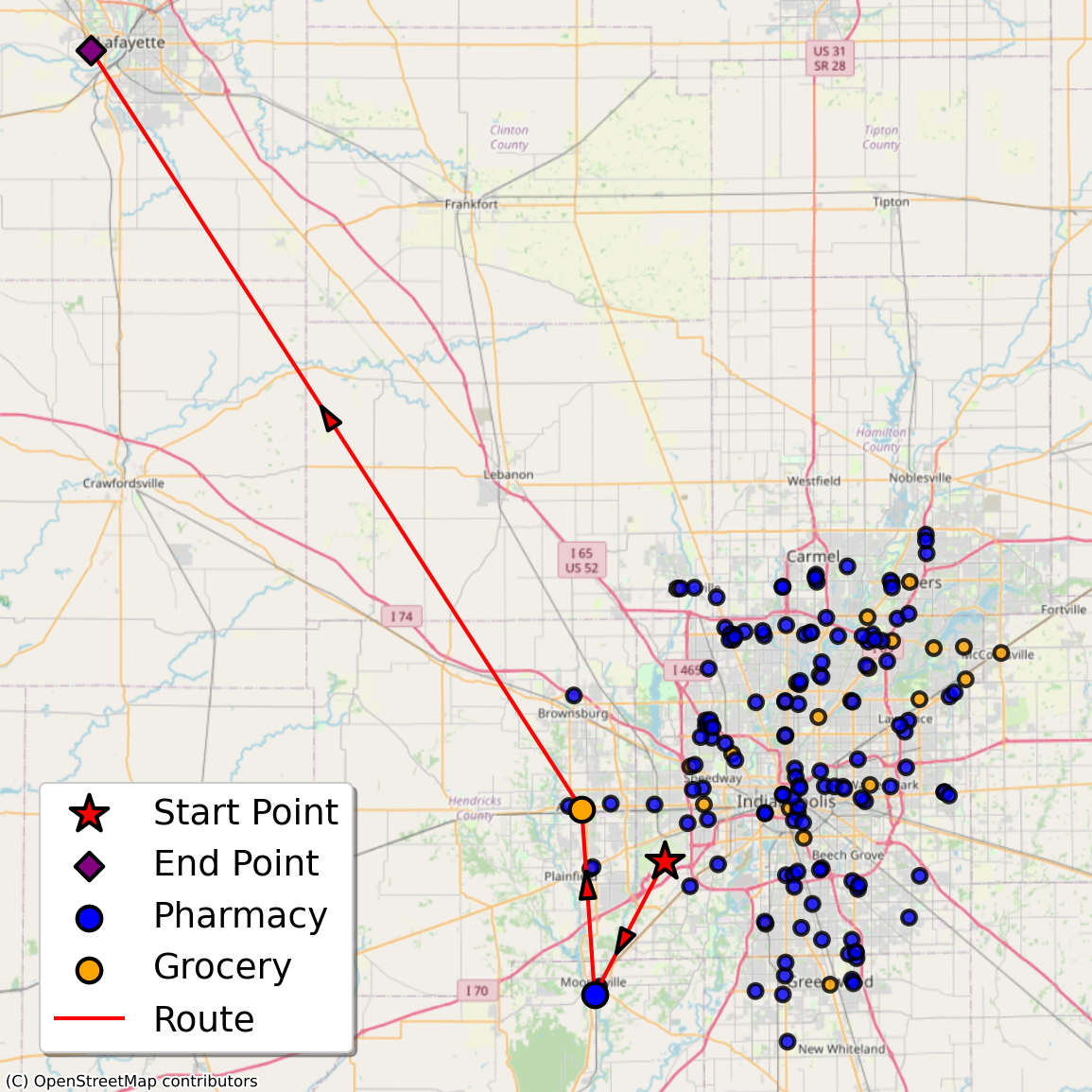}
    }
    \hfill
    \subfloat[Path Planning\label{Fig. Use_Case_2_Path}]{
        \includegraphics[width=0.45\linewidth]{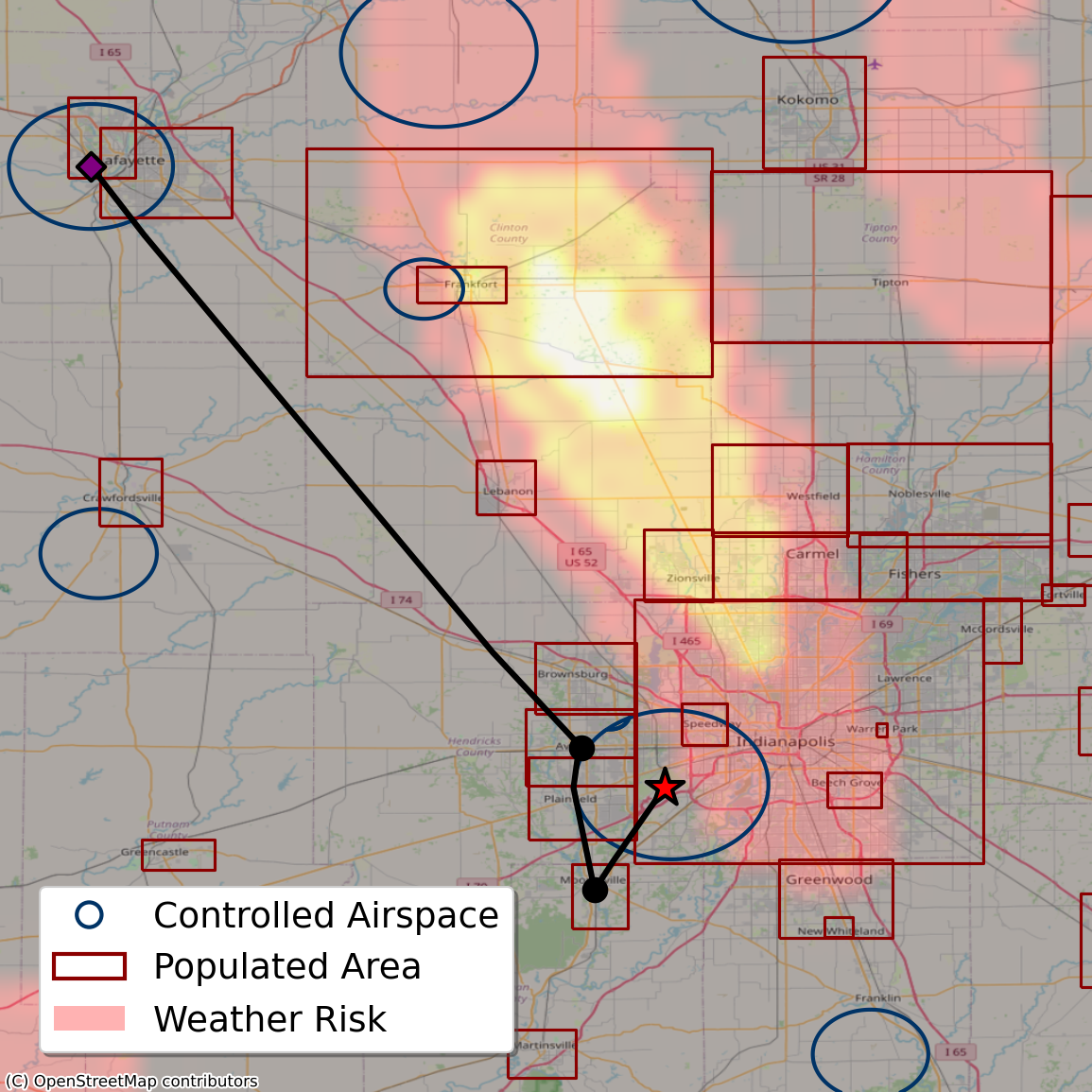}
    }
    \vskip1ex
    \subfloat[Executable Trajectory \label{Fig. Use_Case_2_Traj}]{
        \includegraphics[width=1.0\linewidth]{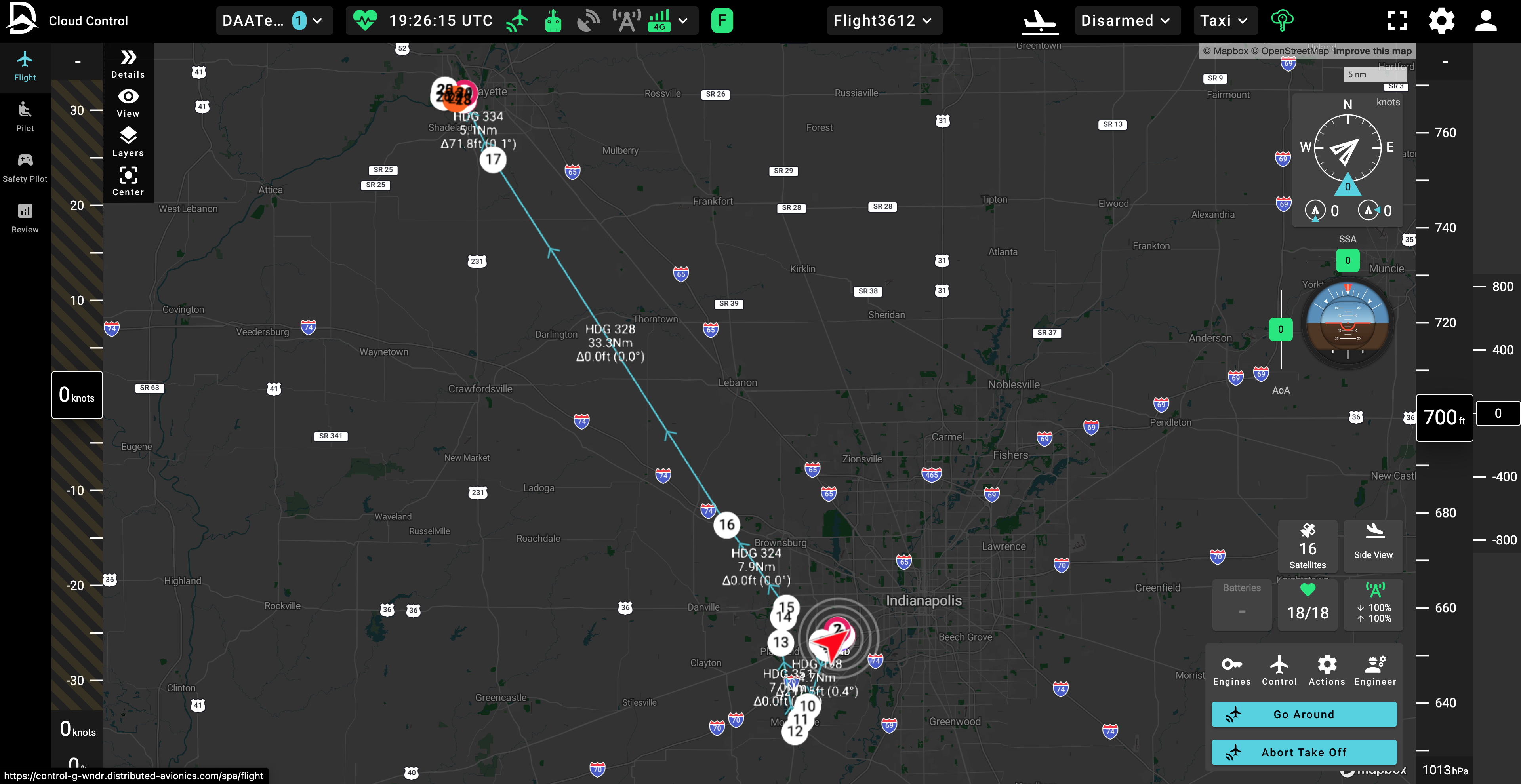}
    }
\end{minipage}
\caption{Use Case 2: Medium-Range Multi-POI Delivery (Sec.~\ref{Sec. Use Case 2}).}
\vspace{-4mm}
\label{Fig. Use_Case_2}
\end{figure*}

\begin{figure*}[t]
\centering
\begin{minipage}[b]{0.5\textwidth}
    \centering
    \subfloat[LLM-as-Parser\label{Fig. Use_Case_3_LLM}]{
        \includegraphics[height=\linewidth]{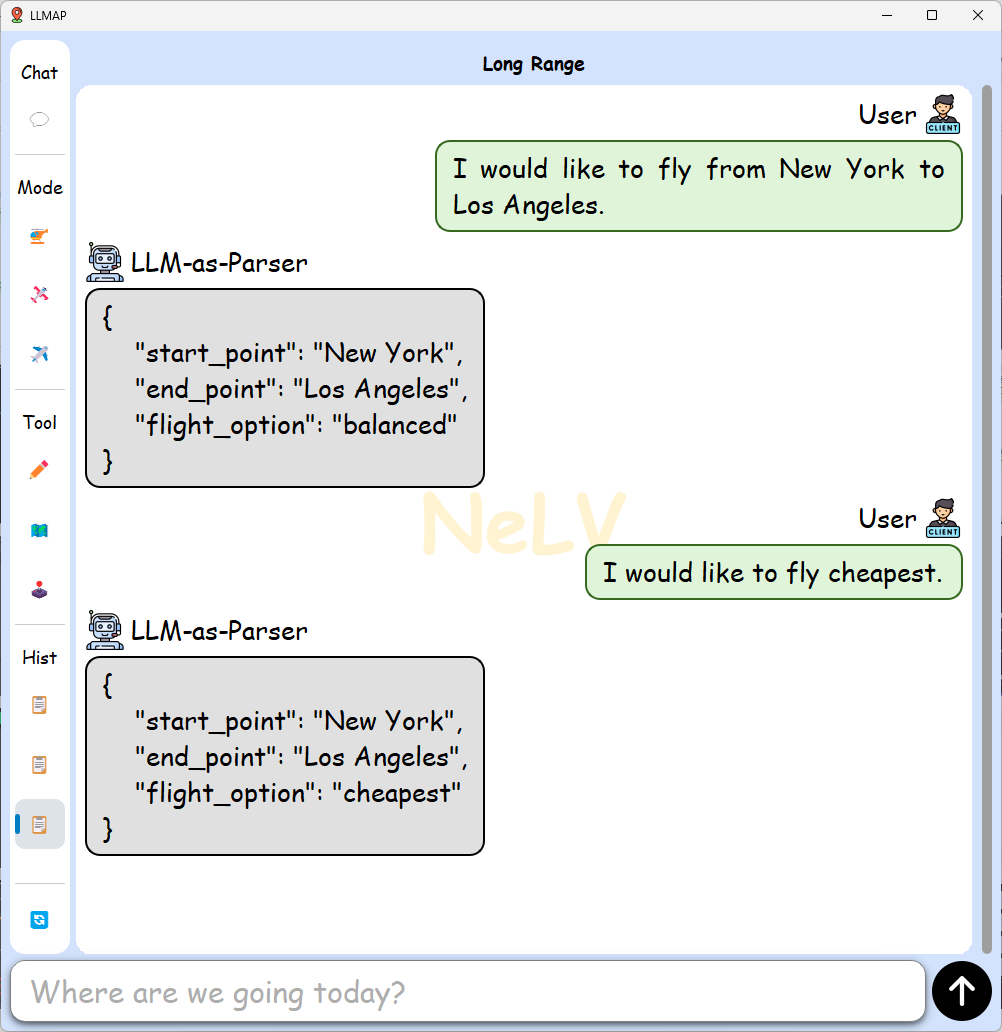}
    }
\end{minipage}
\hfill
\begin{minipage}[b]{0.48\textwidth}
    \subfloat[Route Planning\label{Fig. Use_Case_3_Route}]{
        \includegraphics[width=0.45\linewidth]{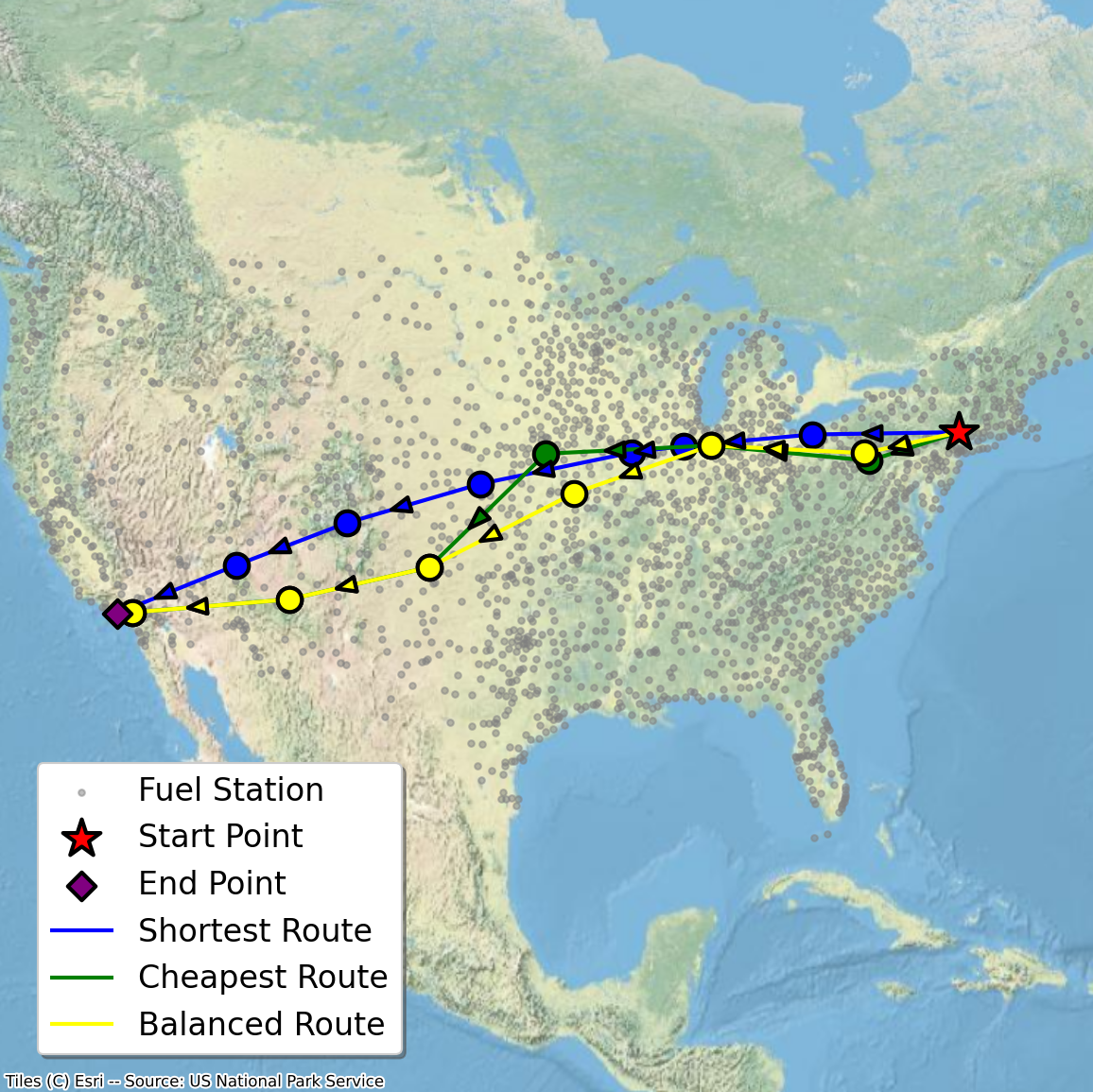}
    }
    \hfill
    \subfloat[Path Planning\label{Fig. Use_Case_3_Path}]{
        \includegraphics[width=0.45\linewidth]{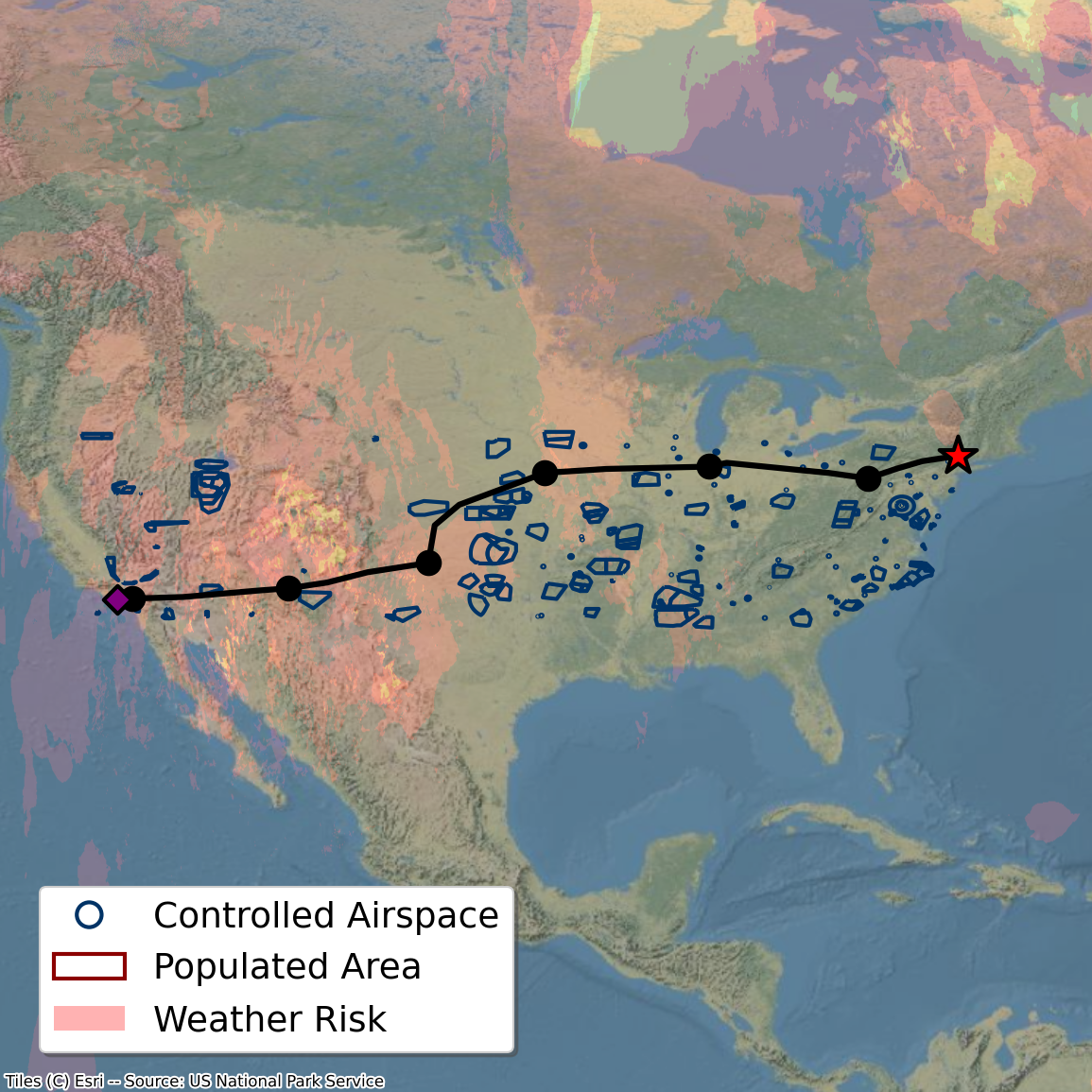}
    }
    \vskip1ex
    \subfloat[Executable Trajectory \label{Fig. Use_Case_3_Traj}]{
        \includegraphics[width=1.0\linewidth]{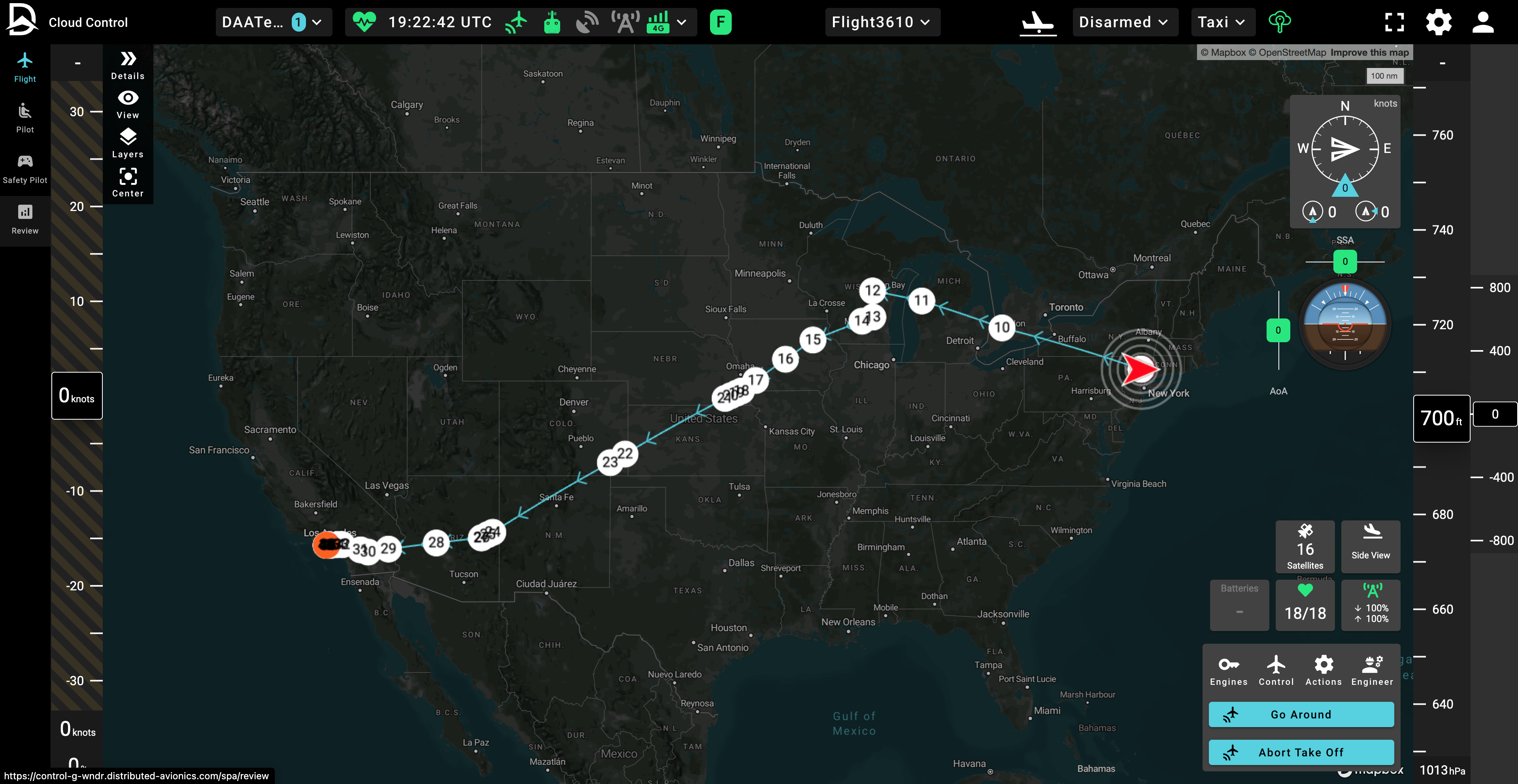}
    }
\end{minipage}
\caption{Use Case 3: Long-Range Multi-Hop Relocation (Sec.~\ref{Sec. Use Case 3}).}
\vspace{-4mm}
\label{Fig. Use_Case_3}
\end{figure*}

We use PyQt5 to develop a graphical user interface (GUI) that functions as the Chat Box, enabling the human pilot to interact with the LLM. The interface integrates both route planning and path planning functionalities, as illustrated in Figures \ref{Fig. Use_Case_1_LLM}, \ref{Fig. Use_Case_2_LLM}, and \ref{Fig. Use_Case_3_LLM}. The GUI consists of three main components: Toolbar, Message Panel, and Input Box. The \textbf{Toolbar} allows the pilot to switch between different LLM modes. From top to bottom, the available options are: Chat (Speech Balloon Emoji), Short Range (Helicopter Emoji), Medium Range (Small Airplane Emoji), Long Range (Airplane Emoji), Plan Route (Pencil Emoji), Plan Path (World Map Emoji), Upload Path (Joystick Emoji), Hist - Short Range (Clipboard Emoji), Medium Range (Clipboard Emoji), Long Range (Clipboard Emoji), and Reset (Counterclockwise Arrows Button Emoji). Pilots employ Chat for general-purpose queries, while the three task modes enable planning for different missions. At any point during the interaction, pilots can activate Plan Route, Plan Path, or Upload Path to perform route planning, path planning, or upload the path to the control platform. In addition, the three Hist functions store past conversations and provide sample tasks. Finally, pilots can reset the conversation to initiate a new mission.

When pilots enter text into the \textbf{Input Box} and click the Send button (Upward Arrow), the text is combined with a predefined system prompt and submitted to the LLM for inference. The system stores conversational context so that pilots can iteratively refine the flight plan. The \textbf{Message Panel} displays all responses, including the LLM’s outputs as well as generated routes and paths. Upon receiving pilot commands, the LLM interprets the input and produces an initial text-based flight plan. Once the pilot accepts this plan, clicking Plan Route triggers the system to generate a route, which is then displayed in the Message Panel. If the route is accepted, the pilot proceeds by clicking Plan Path, prompting the system to produce the corresponding path. Finally, once the path is approved, the pilot clicks Upload Path to send it to the control platform, after which subsequent operations can be performed within the control environment.

\subsection{Control Platform and Real UAV Configurations}

We utilize the Windracers ULTRA UAV \cite{ultra} in conjunction with its integrated control platform, comprising Windracers Mission Control and Windracers Autopilot \cite{mission_control}, to execute flight control and operational missions \cite{jordan2025successful,jordan2025new}. Mission Control constitutes a sophisticated multi-UAV distributed control architecture that provides real-time telemetry feedback regarding aircraft operational status. The platform enables pilots to dynamically modify or reconfigure mission parameters during flight operations; for instance, pilots can manually delete waypoints, prompting autonomous navigation to subsequent destinations, or adjust waypoint coordinates (latitude, longitude, and altitude) through either interactive map manipulation or direct numerical parameter editing. Mission Control additionally supports advanced operational capabilities, including loiter point configurations that enable aircraft to maintain loitering patterns over designated surveillance areas. Furthermore, the platform facilitates simultaneous visualization and control of multiple UAV assets, enabling coordinated monitoring and collaborative mission execution.

The ULTRA \cite{ultra} is a fixed-wing UAV with a maximum take-off mass of 510 kg and a useful payload capacity of 150 kg. The recently updated ULTRA MK2 extends this capability by transporting up to 80 liters of cargo over ranges of over 1,000 km. Powered by two HIRTH F23 engines, each rated at 45 horsepower, and fueled by a standard unleaded petrol/oil mixture, the aircraft achieves a cruise speed of approximately 40 m/s and operates at altitudes up to 4,000 m. Depending on configuration, it sustains autonomous flights for 7–9 hours and is compatible with diverse runway conditions, including paved, grass, gravel, and dirt surfaces. Beyond its physical specifications, the ULTRA platform supports a range of mission profiles: cargo delivery, parachute drops for critical supplies, and aerial survey and detection applications such as wildfire monitoring.

\subsection{Use Case Setup}

We next present three representative use cases of varying complexity, corresponding to short-, medium-, and long-range missions. Furthermore, we examine three specific operational challenges: multi-UAV patrol, multi-POI delivery, and multi-hop relocation, which are evaluated across three distinct datasets and sources, including OpenStreetMap \cite{OpenStreetMap}, the Yelp Open Dataset \cite{yelp-opendataset}, and USA airport fuel prices and specifications from AirNav \cite{airnav}. We aim to demonstrate the scalability of the \Name system through these diverse data sources, which provide geospatial information through varied methodologies and formats. For instance, OpenStreetMap serves as a comprehensive map search service enabling users to query information for any global location, while the Yelp Open Dataset constitutes a preprocessed repository containing extensive POI geographical coordinates and user review data. Additionally, users can deploy the \Name system on proprietary datasets or commercial mapping services such as Google Maps, demonstrating our system's adaptability across different data infrastructures.

In our experimental implementation, path planning draws on airspace information, population density data, and weather forecasts. Airspace information is obtained from OpenAIP \cite{openaip}, which for the continental United States comprises 3,244 published volumes, including Class B, C, D, and E airspace as well as restricted, prohibited, and danger areas. Volumes are retained according to their vertical limits relative to the planned cruise altitude, so that short- and medium-range missions consider airspace extending below 2,000 ft while long-range missions consider airspace extending above 15,000 ft. Population density \cite{OpenStreetMap} enters as ground risk, which the system evaluates against city boundaries and urban area geometries and disables for long-range flights operating at high altitudes. Weather forecast data is sourced from the High-Resolution Rapid Refresh (HRRR) model stored in the Herbie dataset \cite{herbie} at 250 mb for long-range flights and 950 mb for short-range operations, from which cloud, temperature, humidity, vertical velocity, Convective Available Potential Energy (CAPE), wind shear, and horizontal wind fields are combined into the Simplified Forecast Icing Potential (SFIP) and the Bulk Richardson Number (BRN) and integrated along each path segment. Forecasts are retrieved at run time, and we pin the forecast date in our released configuration so that the reported results remain reproducible. For trajectory generation, each take-off and landing airport is described by its runway threshold position, true alignment, field elevation, and traffic-pattern direction, all of which are published for every public-use runway end in the OurAirports database \cite{ourairports}, and from which Algorithm \ref{Alg. Circuit} produces the corresponding circuit. All reported results use $w_0 = 10$, $w_1 = 1$, $w_2 = 0.05$, and $w_3 = 0.1$ in Eq.~(\ref{Eq. Path Planning Instantiated}), with the constraint weight set an order of magnitude above the cost weights so that no cost saving offsets a violation. The remaining planner parameters are listed in the released implementation.

\subsection{Use Case 1: Short-Range Multi-UAV Patrol}
\label{Sec. Use Case 1}

\noindent\textbf{Implementation Details:} We utilize OpenStreetMap \cite{OpenStreetMap} as the primary source of geospatial information. OpenStreetMap constitutes a freely accessible, open-source mapping service that enables pilots to query geographical elements through keyword-based searches, including amenity classifications (e.g., education, healthcare, transportation), building categories (e.g., sports, automotive, storage), commercial establishments (e.g., food, beauty, clothing), and various other geographic features. The platform supports multiple query methodologies for identifying objects within target regions, encompassing bounding box queries, administrative boundary searches, distance-based range queries, and polygonal area selections. In our implementation, we establish the map center at Purdue University Airport and examine forested areas within a 5 km radius to facilitate a multi-UAV patrol mission focused on wildfire detection.

Figure \ref{Fig. Use_Case_1_LLM} demonstrates the interactive dialogue between the pilot and the LLM, where initial mission parameters are specified, including the origin point, the detection objective, the surveillance range, and the size of the UAV fleet. In the initial dialogue exchange, the LLM successfully identifies the mission parameters; however, given the absence of explicit UAV quantity specification, the system defaults to a single-UAV configuration. Following the pilot's specification of five UAVs, the LLM dynamically updates the mission plan accordingly. Upon pilot confirmation of the flight plan, the route planning algorithm is initiated. For multi-UAV route optimization, we employ OR-Tools \cite{ortools_routing}, with the resulting routes illustrated in Figure \ref{Fig. Use_Case_1_Route}. The forested terrain adjacent to Purdue University is partitioned into five discrete sectors, each assigned a designated UAV route. The corresponding flight paths are depicted in Figure \ref{Fig. Use_Case_1_Path}; notably, in this short-range operational scenario, the paths coincide with the routes due to the limited operational range and altitude constraints, which render weather, airspace, and terrain-based risks negligible. Figure \ref{Fig. Use_Case_1_Traj} illustrates the complete executable trajectory, including take-off and landing patterns, as well as flight paths to execute the mission. We demonstrate two distinct operational patterns utilizing different runway orientations: the take-off pattern aligned in a northeast-southwest direction and landing oriented in a nearly east-west configuration. These differentiated runway approaches prevent runway incursions when multiple UAVs execute concurrent missions at the same airport facility.

\subsection{Use Case 2: Medium-Range Multi-POI Delivery}
\label{Sec. Use Case 2}

\noindent\textbf{Implementation Details:} We utilize the Yelp Open Dataset \cite{yelp-opendataset} as the primary source of POI information for mission planning, encompassing geographical coordinates, establishment ratings, review counts, and operational hours. In our implementation, we consider a use case wherein a UAV departs from Indianapolis Airport and navigates to Purdue University Airport, with intermediate stops at a pharmacy and a supermarket for supply procurement. The mission objective is to maximize the quality of these two locations (i.e., ratings and review counts) while simultaneously minimizing the total route distance. At each POI, the system automatically establishes a loiter point where the UAV executes circular flight patterns at approximately 300 meters Above Ground Level (AGL) to simulate cargo operations. In practical deployment scenarios, this configuration can be dynamically adapted based on operational requirements, such as executing precision landings for ground-based cargo pickup, performing aerial cargo drops through automated release mechanisms, or maintaining extended surveillance patterns for reconnaissance missions.

In Figure \ref{Fig. Use_Case_2_LLM}, the pilot specifies a flight trajectory from Indianapolis to Purdue University with an intermediate stop at a pharmacy. The \Name system accurately recognizes the start and end points while correctly classifying the pharmacy as a POI. During this use case, the pilot dynamically adds an additional POI by requesting a visit to a supermarket. Following the LLM's acknowledgment and the pilot's confirmation of mission parameters, the pilot utilizes the interface tools to execute Algorithm \ref{Alg. Route} for route optimization, with results presented in Figure \ref{Fig. Use_Case_2_Route}. Subsequently, Figure \ref{Fig. Use_Case_2_Path} illustrates the path transformation from the computed route, which incorporates meteorological hazards, terrain-based risks, and airspace restrictions. In Figure \ref{Fig. Use_Case_2_Path}, the red boundaries delineate municipal and township limits, while the blue circles represent controlled airspace requiring air traffic control clearance. Additionally, a superimposed heatmap visualizes weather-related risk distributions across the operational area. The path planning algorithm enables weight adjustment to prioritize specific cost functions over others; in this particular use case, we configure the system to prioritize path distance optimization over weather and ground risk avoidance, while the airspace constraint retains the highest weight.

\subsection{Use Case 3: Long-Range Multi-Hop Relocation}
\label{Sec. Use Case 3}

\noindent\textbf{Implementation Details:} We utilize airport data obtained from AirNav \cite{airnav}, encompassing locations, identifiers, and fuel pricing for all airports across the United States. Fuel prices exhibit substantial variation among different airport locations. Based on data acquired from AirNav in March 2025, the continental United States contains 2,577 airports, of which 2,076 provide fuel pricing information. Aviation fuel prices exhibit substantial variation across locations, ranging from a minimum of 0.74 USD/L at Clintonville Municipal Airport to a maximum of 2.77 USD/L at Kodiak Airport. Consequently, when UAVs conduct long-range multi-hop relocation missions, the trade-off between fuel costs at different airports and flight distances must be carefully considered. In our configuration, the ULTRA UAV features a fuel tank capacity of 80 L with a fuel consumption rate of 10.95 km/L, yielding a maximum flight range of approximately 876 km. Note that fuel consumption varies with mass and weather conditions. Due to airport traffic control constraints and varying operational complexity, estimating the actual fuel consumption during take-off and landing procedures at individual airports presents significant challenges. Therefore, we establish a standardized fuel overhead of approximately 10 liters per airport operation (take-off and landing combined), equivalent to approximately one hour of cruise flight consumption. This establishes a critical path planning constraint wherein edges between airports cannot exceed 876 km, enabling the construction of a graph network for route planning optimization. All restricted airspace between 15,000 and 40,000 feet are considered given the cruise altitude of approximately 30,000 feet. Meteorological hazards including thunderstorm activity, turbulence zones, and icing potential are incorporated into the path planning algorithm.

In Figure \ref{Fig. Use_Case_3_LLM}, the pilot specifies a transcontinental flight from New York to Los Angeles. The start and end points are accurately recognized by the system. Given that long-range flights necessitate multiple refueling stops, \Name incorporates a flight option feature enabling pilot selection among optimization strategies. By default, the system employs a \textit{balanced} configuration, which generates routes that optimize the trade-off between total route distance and cumulative cost. Figure \ref{Fig. Use_Case_3_Route} presents all feasible route alternatives. In this instance, the pilot selects the \textit{cheapest} option, prompting the system to reconfigure the flight strategy and select the corresponding cost-optimized route. The resulting path derived from this route selection is illustrated in Figure \ref{Fig. Use_Case_3_Path}. Notably, given the cruise altitude of approximately 30,000 feet for long-range operations, terrain-based risks are rendered negligible, and the path planning algorithm considers airspace restrictions and meteorological hazards.

\section{Quantitative Evaluation}
\label{Sec. Eval}

\subsection{LLM-as-Parser Accuracy}
\label{Sec. Eval Parser}

\noindent\textbf{Setup:} LLM-as-Parser maps a conversation to a JSON object whose fields are fixed by the mission type, and retains previously established values when the pilot revises the mission in a later turn. We evaluate this capability on the Schema-Guided Dialogue benchmark \cite{rastogi2020towards}, a public corpus of task-oriented dialogues in which every user turn is annotated with the fields the user has specified. We sample 300 dialogues from the test split, comprising 3391 user turns and 10921 field values, and supply each extractor with the schema and the dialogue history up to that turn. Accuracy is reported on three kinds of field a UAV mission may carry, namely where the flight begins and ends, when it has to fly, and what it is looking for. Three non-neural extractors are compared against eight LLMs spanning four families and five parameter counts. Keyword matching scans each turn for the field name and its aliases, the second baseline assigns each spaCy named entity to the field its type admits, and the third matches each turn against a lexicon of the values the training split records for that field. All three propagate state across turns. Every LLM is served with vLLM on a single H100 and decoded greedily under a 512-token budget, and latency is the median over 100 single-request turns.

Table \ref{Table Eval Parser} reports parsing accuracy by field kind, schema validity, and the latency of a single parse. Every LLM evaluated exceeds the strongest non-neural extractor by at least 28.6 accuracy points, and the margin widens as the value space of a field becomes less constrained. Keyword matching recovers no value whatsoever, since it can only match the enumerated values a schema provides, and the entity recogniser and the value lexicon recover roughly half of the \textit{Location} and \textit{Time} values through surface patterns. All three baselines fail on \textit{Name} fields, whose values are free text drawn from no fixed vocabulary, and it is here that the advantage of the LLMs is largest. This is also the field kind on which a pilot's instruction depends, since an airport, a landmark, or a point of interest is referred to in unconstrained language. Within the Qwen3.5 family, accuracy increases monotonically with parameter count from 0.8B to 27B. We further validate two properties that may affect deployment. First, \textit{Valid} is high across every LLM evaluated, remaining above 90\% for the weakest and above 99\% for most. Second, \textit{Latency} is not a constraint, staying below one second at every scale and accounting for under 10\% of the elapsed time of a medium- or long-range mission in Table \ref{Table Latency}.

\begin{table}[t]
    \centering
    \caption{Parsing Accuracy of LLM-as-Parser on the Schema-Guided Dialogue Benchmark. The number of evaluated field values is given in parentheses, and $\pm$ denotes a 95\% confidence interval.}
    \label{Table Eval Parser}
    \resizebox{\linewidth}{!}{
    \begin{tabular}{lccc|a|cc}
    \toprule
    \textbf{Method} & \textbf{Location} & \textbf{Time} & \textbf{Name} & \textbf{Avg.} & \textbf{Valid} & \textbf{Latency} \\
     & (2805) & (2626) & (1227) & (6658) & (\%) & (ms) \\
    \midrule
    Keyword matching & 0.0 \com{0.0} & 0.0 \com{0.0} & 0.0 \com{0.0} & 0.0 \com{0.0} & 100.0 \com{0.0} & 0 \com{0} \\
    spaCy NER + rules & 40.9 \com{4.3} & 41.3 \com{4.1} & 15.9 \com{4.2} & 36.5 \com{2.7} & 100.0 \com{0.0} & 4 \com{0} \\
    Value lexicon & 45.1 \com{4.8} & 53.4 \com{4.3} & 5.6 \com{2.4} & 41.1 \com{2.9} & 100.0 \com{0.0} & 1 \com{0} \\
    \midrule
    Qwen3.5-0.8B & 74.3 \com{3.4} & 64.1 \com{3.6} & 71.3 \com{4.9} & 69.7 \com{2.6} & 99.4 \com{0.2} & 181 \com{13} \\
    Qwen3.5-2B & 83.4 \com{2.8} & 69.6 \com{3.4} & 84.2 \com{3.6} & 78.1 \com{2.2} & 99.4 \com{0.3} & 172 \com{17} \\
    Qwen3.5-4B & 87.4 \com{2.1} & 76.5 \com{3.3} & 73.8 \com{4.6} & 80.6 \com{2.1} & 99.9 \com{0.1} & 285 \com{41} \\
    Qwen3.5-9B & 88.7 \com{2.2} & 80.4 \com{2.9} & 69.5 \com{4.6} & 81.9 \com{1.8} & 99.9 \com{0.2} & 412 \com{64} \\
    Qwen3.5-27B & 90.5 \com{1.9} & 85.0 \com{2.7} & 80.4 \com{3.8} & 86.5 \com{1.6} & 99.9 \com{0.1} & 743 \com{127} \\
    Phi-4-mini-instruct & 80.7 \com{2.8} & 67.4 \com{3.4} & 59.0 \com{5.4} & 71.5 \com{2.2} & 90.1 \com{1.4} & 204 \com{31} \\
    Gemma-4-E4B-it & 89.1 \com{2.3} & 78.0 \com{3.1} & 48.7 \com{5.4} & 77.2 \com{2.2} & 99.9 \com{0.1} & 222 \com{35} \\
    Llama-3.1-8B-Instruct & 88.1 \com{2.4} & 76.0 \com{3.5} & 67.7 \com{4.7} & 79.6 \com{2.3} & 92.2 \com{1.1} & 524 \com{47} \\
    \bottomrule
    \end{tabular}
    }
\vspace{-4mm}
\end{table}

\subsection{End-to-End Planning Success}
\label{Sec. Eval Planning}

\noindent\textbf{Setup:} We sample 1000 random pairs of airports from the AirNav dataset within Indiana and 1000 more across the continental United States, and run the whole pipeline on each pair. Medium-range and long-range missions present different challenges. A medium-range mission flies a single edge at low altitude, where one tank covers the distance several times over and fuel never binds, and where the difficulty lies in the density of controlled volumes at that altitude and in the proximity of the airports, which leaves a path little lateral room. A long-range mission is constrained instead by fuel, since the aircraft has to reach a refuelling stop before the tank runs dry, and route planning sizes each edge close to the range of the aircraft, which leaves path planning almost nothing to spend on a deviation. The cruise altitude also determines what has to be avoided, since the volumes that reach a long-range cruise are the prohibited and restricted ones alone.

Table \ref{Table Path Bench} reports what each stage returns and what its plan holds. Route planning and the control platform succeed on every mission, which follows from the way each is constructed. Route planning searches the graph of reachable edges with Dijkstra's algorithm, so it returns a route whenever the two endpoints are connected, and that route is optimal under whichever flight option the pilot selects, which is the balanced option here. The control platform derives the circuits in closed form from the runway geometry, a construction prescribed by traffic pattern convention and kept deterministic for safety. The stage that falls short is path planning, for a different reason in each class. Of the 1000 medium-range missions, 26 enter restricted airspace, because we configure the search with two free waypoints per edge, which makes a candidate path three straight segments and rarely enough to bend around corridors as narrow as the ones separating neighbouring Indiana airports. Of the 1000 long-range missions, 23 fail, 17 by entering restricted airspace, seven by a turn the airframe cannot hold, and one by exhausting its fuel on an edge route planning had sized near the range of the aircraft, two of them on two counts at once. These are the missions on which PSO converges to a local optimum and the search stops short of a feasible path. The planner reports the constraints the returned path still violates, so a plan of this kind is flagged as it is produced, and the system either searches again or passes it to the pilot for adjustment through the control platform before the UAV flies.

\begin{table}[t]
    \centering
    \caption{Success Rate of Our \Name System over 1000 Random Airport Pairs in Each Mission Class.}
    \label{Table Path Bench}
    \footnotesize
    \resizebox{\linewidth}{!}{
    \begin{tabular}{lcc}
    \toprule
    \textbf{Metric} & \textbf{Medium-Range} & \textbf{Long-Range} \\
     & (Indiana) & (United States) \\
    \midrule
    \textbf{Route Success} (\%) $\uparrow$ & 100.0 \com{0.0} & 100.0 \com{0.0} \\
    \quad Fuel Exhausted (\%) $\downarrow$ & 0.0 \com{0.0} & 0.0 \com{0.0} \\
    \quad Fuel Consumption (L/km) $\downarrow$ & - & 0.115 \com{0.001} \\
    \quad Fuel Price (USD/L) $\downarrow$ & - & 1.076 \com{0.015} \\
    \quad Fuel Cost (USD/km) $\downarrow$ & - & 0.124 \com{0.002} \\
    \textbf{Path Success} (\%) $\uparrow$ & 97.4 \com{1.0} & 97.7 \com{1.0} \\
    \quad Airspace Violation (\%) $\downarrow$ & 2.6 \com{0.9} & 1.7 \com{0.8} \\
    \quad Fuel Exhausted (\%) $\downarrow$ & 0.0 \com{0.0} & 0.1 \com{0.2} \\
    \quad Turn-Angle Violation (\%) $\downarrow$ & 0.0 \com{0.0} & 0.7 \com{0.6} \\
    \textbf{Trajectory Success} (\%) $\uparrow$ & 100.0 \com{0.0} & 100.0 \com{0.0} \\
    \midrule
    \textbf{Mission Success} (\%) $\uparrow$ & 97.4 \com{1.0} & 97.7 \com{1.0} \\
    \bottomrule
    \end{tabular}
    }
\vspace{-4mm}
\end{table}

\subsection{System-Level Latency Analysis}
\label{Sec. Eval Latency}

Table \ref{Table Latency} reports the latency of each component of \Name. LLM-as-Parser spends its time on a single forward pass of the LLM, Route Planner on the search over the subgraph of candidate nodes, and Control Platform on the closed-form geodetic transformations that turn the path into an executable trajectory. Each of these completes in under a second. Path Planner dominates, and its latency is the latency of the slowest edge of the route, since the edges share no state and Algorithm \ref{Alg. Path} searches them in parallel. What makes a single edge expensive is the cost of one objective evaluation of Eq. (\ref{Eq. Path Planning Instantiated}). Each evaluation integrates the service data along the whole length of a candidate path, testing every segment against the airspace volumes it enters, the forecast grid cells it crosses, and the populated areas it overflies, and the search repeats this for every candidate path it generates. The latency of this component therefore scales with the number of terms the cost and constraint functions carry. The medium-range mission activates the most of them, since it evaluates ground risks over populated areas in addition to the weather risks and airspace restrictions that the other missions already impose.
    
\begin{table}[t]
    \centering
    \caption{Per-Component Latency of Our \Name System in Seconds over Ten Replays of Each Use Case.}
    \label{Table Latency}
    \footnotesize
    \resizebox{\linewidth}{!}{
    \begin{tabular}{lcccccc}
    \toprule
     & \multicolumn{5}{c}{\textit{\textbf{Latency (s)}}}  \\
    \cmidrule(lr){2-6}
    \textbf{Use Case} & \textbf{LLM} & \textbf{Route} & \textbf{Path} & \textbf{Trajectory} & \textbf{Total} \\
    \midrule
    Short-Range & 0.16 \com{0.00} & 0.03 \com{0.01} & - & $<$0.01 & 0.19 \\
    Medium-Range & 0.14 \com{0.00} & 0.07 \com{0.02} & 1.40 \com{0.09} & $<$0.01 & 1.61 \\
    Long-Range & 0.13 \com{0.00} & 0.75 \com{0.15} & 1.36 \com{0.08} & $<$0.01 & 2.24 \\
    \bottomrule
    \end{tabular}
    }
\vspace{-4mm}
\end{table}

\section{Roadmap and Future Directions}
\label{Sec. Future Direction}

\begin{figure}[t]
    \centering
    \includegraphics[width=1\linewidth]{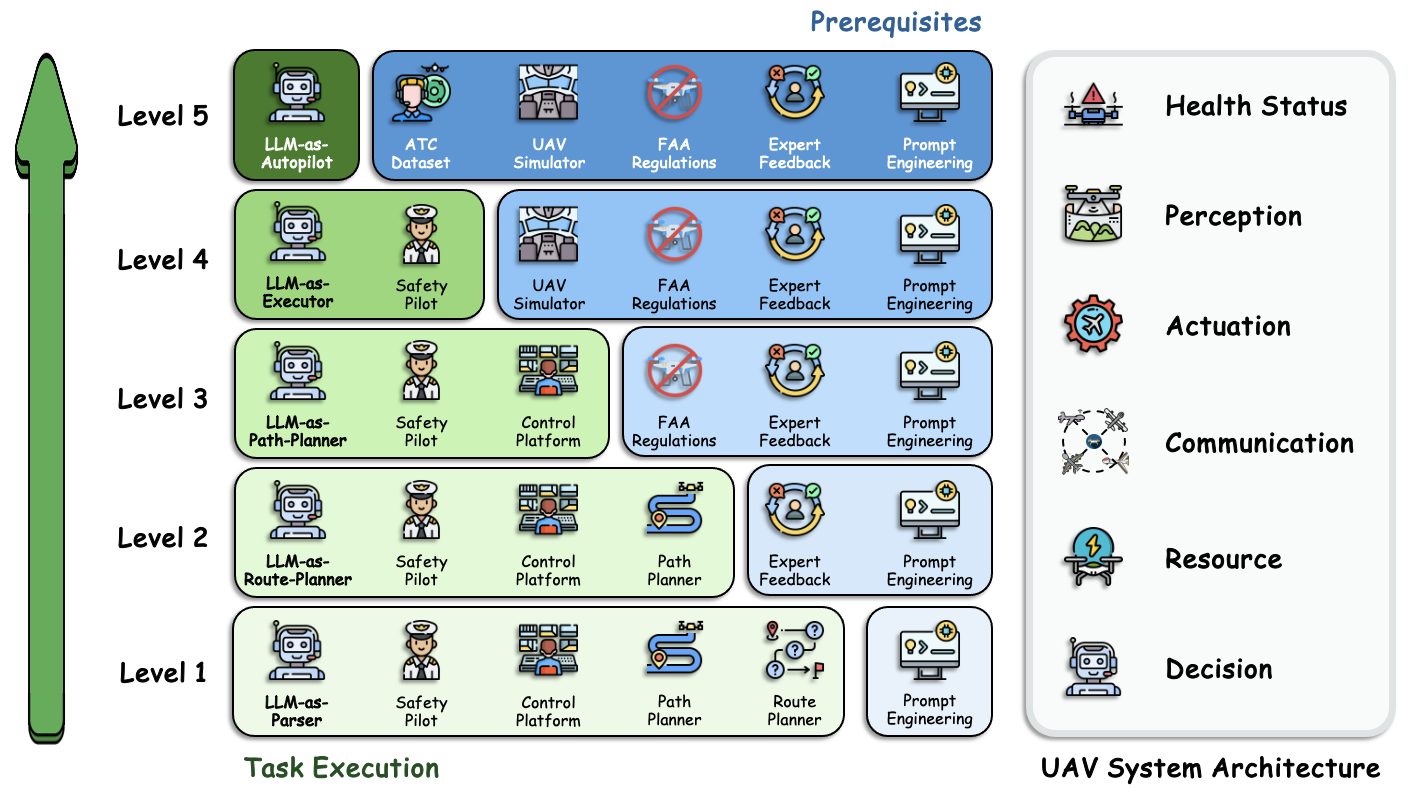}
\vspace{-4mm}
    \caption{Evolutionary Roadmap of Our \Name System. \textbf{Green Box – Task Execution:} Components required for \Name system operation, demonstrating progressive integration where higher automation levels reduce component complexity while achieving enhanced system integration, evolving from Level 1 (LLM-as-Parser) to Level 5 (LLM-as-Autopilot). \textbf{Blue Box – Prerequisites:} Infrastructure and knowledge requirements that expand with each automation level to support increasingly sophisticated LLM reasoning capabilities. \textbf{Gray Box – UAV System Architecture:} Essential UAV sensing, communication, and operational interface components that provide foundational support across all automation levels.}
\vspace{-4mm}
    \label{Fig. Roadmap}
\end{figure}

The \Name architecture comprises five discrete components to accommodate the evolving capabilities of general-purpose LLMs in UAV applications. Present-day LLMs require specialized integration with domain-specific knowledge including airspace regulations, aeronautical mapping data, UAV control systems, operational protocols, and swarm intelligence. As UAV-specific LLMs mature on comprehensive UAV datasets, this modular design allows components to consolidate. As illustrated in Figure \ref{Fig. Roadmap}, this roadmap delineates the information requirements, decision-making capabilities, and technical prerequisites for LLMs at each automation level, ultimately leading to fully autonomous UAV mission planning and execution.

\subsection{Level 1: LLM-as-Parser}

\noindent\textbf{Prerequisite:} Prompt engineering.

\noindent\textbf{Task Execution:} Route planner (Algorithm \ref{Alg. Route}), path planner (Algorithm \ref{Alg. Path}), control platform (Algorithm \ref{Alg. Circuit}), and safety pilot.

\noindent\textbf{LLM's Output:} Start point, end point, and POIs.

\noindent\textbf{Responsibility:} The \Name system currently is Level 1, which employs LLM as a parser to extract information from human pilot instructions. General-purpose LLMs adequately perform this function, as they require no specialized UAV knowledge but focus exclusively on token extraction. In more complex operational scenarios, LLMs must extract sophisticated constraints, exemplified when a human pilot requests ``please ensure arrival before the airport closes,'' which contains the time constraint ``airport closing.'' In such instances, the LLM need not comprehend the semantic meaning of ``airport closing'' but simply process it as a predefined identifier for subsequent evaluation during route planning (as implemented in Algorithm \ref{Alg. LLM} - Line 4).

\noindent\textbf{Challenges:} Since we do not require the LLM to engage in decision-making, it remains incapable of managing operations beyond predetermined keywords. For LLM-as-Parser, we must predefine the specific information to be extracted in the system prompt (e.g., names of the start point and the end point), so it fails to comprehend non-predefined information, such as ``this is an urgent medicine delivery," and does not incorporate this requirement in the route planning process. A further limitation is that LLM-as-Parser does not integrate contextual information to facilitate more sophisticated inference. For example, in relocation scenarios where certain airports experience snowstorms, aviation best practices dictate avoiding these locations, but LLM-as-Parser cannot eliminate these airports during the route planning phase. Human pilots must then exclude them manually rather than through natural language, increasing their operational burden.

\subsection{Level 2: LLM-as-Route-Planner}

\noindent\textbf{Prerequisites:} Prompt engineering and expert feedback.

\noindent\textbf{Task Execution:} Path planner (Algorithm \ref{Alg. Path}), control platform (Algorithm \ref{Alg. Circuit}), and safety pilot.

\noindent\textbf{LLM's Output:} Route (i.e., a sequence of nodes).

\noindent\textbf{Responsibility:} Level 2 implements LLM-as-Route-Planner, which requires comprehensive knowledge of the operational environment, specifically the nodes, including their geolocations, operational status, and mission-critical parameters such as fuel price. Beyond knowledge acquisition, it must interpret the implicit objectives underlying human instructions and compute viable routes that satisfy them. Given ``I want to fly a UAV to a nearby airport for refueling, considering the most economical flight path," it must evaluate proximate airports, their geospatial positions, fuel expenditure, consumption, and wind. Given ``proceed as quickly as possible, as I need to use it soon," it must identify the temporally optimal route.

\noindent\textbf{Challenges:} LLM-as-Route-Planner requires access to comprehensive geospatial databases of mission-relevant nodes, whose density varies significantly with the mission and the region; according to the Yelp Open Dataset \cite{yelp-opendataset}, Indianapolis contains 54 pharmacies and 234 supermarkets. It must then compute over this information, such as calculating flight duration and ranking airports according to relevant criteria. It must further understand pilot preferences, both explicit and implicit. Explicit preferences are directly articulated within the pilot instructions, including economic efficiency, temporal urgency, or spatial proximity. Implicit preferences are derived from human experiential judgments, which could potentially be addressed through reinforcement learning from human feedback (RLHF) \cite{ouyang2022training, sun2024optimizing, huang2024trustworthy}. These capabilities might be achieved through specialized training or fine-tuning, or through integration with auxiliary APIs.

\subsection{Level 3: LLM-as-Path-Planner}

\noindent\textbf{Prerequisites:} Prompt engineering, expert feedback, airspace information.

\noindent\textbf{Task Execution:} Control platform (Algorithm \ref{Alg. Circuit}) and safety pilot.

\noindent\textbf{LLM's Output:} Path (i.e., a sequence of waypoints).

\noindent\textbf{Responsibility:} Level 3 implements LLM-as-Path-Planner for waypoint determination, which requires significantly more information than Level 2, as it must provide waypoints beyond the nodes specified in the route. While route planning may be considered a common-sense node selection process, waypoint determination demands more extensive aviation domain knowledge, including densely populated areas, airspace classifications, and air traffic density. Like LLM-as-Route-Planner, it must comprehend human pilot preferences, not only regarding node selection but also path features such as flight times and flight paths. A further distinction is that waypoints are three-dimensional, so LLM-as-Path-Planner must determine flight altitude as well as geographical position, adhering to FAA regulations while accommodating human pilot preferences.

\noindent\textbf{Challenges:} The distinctive challenges of path planning involve complying with FAA regulations and planning appropriate flight altitudes. LLMs require profound comprehension of FAA regulations, particularly considering that multiple factors invoke different rules, including UAV type, dimensions, weight, flight timing (e.g., nighttime operations), and flight velocity. The current \Name system requires human pilots to manually specify the UAV type to ensure compliance. LLM-as-Path-Planner must instead select the applicable regulations from the pilot's language input or from system background information. Flight altitude presents a second challenge. Although take-off, landing, and mission execution traffic patterns are designed within the control platform, path altitude during transit affects fuel efficiency, airspace management, weather exposure, and overall flight safety.

\subsection{Level 4: LLM-as-Executor}

\noindent\textbf{Prerequisite:} Prompt engineering, expert feedback, airspace information, and UAV simulator.

\noindent\textbf{Task Execution:} Safety pilot.

\noindent\textbf{LLM's Output:} Trajectory (i.e., a sequence of waypoints including take-off, landing, and mission-specific flight patterns).

\noindent\textbf{Responsibility:} Level 4 further encompasses the functionality of the cloud platform, employing LLM-as-Executor to plan executable trajectories and \textit{remotely control} UAV flight operations. The control platform extends beyond path planning to flight patterns, which is particularly crucial for fixed-wing UAVs that require specific patterns for take-off, landing, and mission execution. Fixed-wing UAVs generally adhere to standard left-hand traffic patterns, so LLM-as-Executor must design patterns from airport specifications. It must also design for mission type. In a mission observing wild yak populations, multi-rotor UAVs can hover, so a three-dimensional coordinate position suffices, whereas fixed-wing UAVs cannot hover and require circular patterns for continuous observation of the same location.

\noindent\textbf{Challenges:} The primary challenge involves extensive knowledge of UAVs, particularly the traffic patterns of different UAV types, which correlate strongly with type and size. LLM-as-Executor must respect UAV-inherent limitations; for example, it cannot command vertical take-off and landing for fixed-wing UAVs, as this would result in insufficient lift and subsequent crash. Standard take-off and landing patterns also vary significantly with size and weight. Mission-related traffic patterns constitute a second challenge, with substantial variation between mission types, including circular flight (e.g., search and rescue), airdrop operations (e.g., medication delivery), and grid coverage (e.g., agricultural spraying). These present greater difficulty than take-off and landing procedures, since airport information is finite and can be collected in advance for LLM-as-Executor training, whereas missions such as search and rescue following natural disasters remain unpredictable.

\subsection{Level 5: LLM-as-Autopilot}

\noindent\textbf{Prerequisite:} Prompt engineering, expert feedback, airspace information, UAV simulator, and ATC datasets.

\noindent\textbf{Task Execution:} LLM-as-Autopilot only.

\noindent\textbf{LLM's Output:} UAV actions and interpretations (to enable human pilot intervention when necessary).

\noindent\textbf{Responsibility:} The final level is LLM-as-Autopilot, where the LLM autonomously controls the UAV to complete missions upon receiving human pilot instructions without further intervention. This does not imply that the LLM directly manipulates the UAV's control surfaces. Rather, it autonomously utilizes integrated APIs for operation, including map service, air traffic, and control interfaces. The emphasis is on autonomous decision-making from current situational assessment, comparable to a qualified human pilot. It must also maintain communication with human pilots, reporting and requesting assistance or control transfer when operational conditions become complex, and share information with other LLM-as-Autopilots. In disaster search and rescue, multiple UAVs can share search area information in real time and coordinate subsequent zone allocations.

\noindent\textbf{Challenges:} The fundamental challenge of LLM-as-Autopilot involves simulating human pilot decision-making, requiring comprehensive understanding of all UAV systems, communication protocols with supervising human pilots, and information exchange with other LLM-as-Autopilots. A more critical challenge emerges when the UAV encounters unpredictable contingencies. After a bird strike, LLM-as-Autopilot must first stabilize the aircraft, then assess all systems, request human pilot intervention if warranted, and alert other UAVs to the avian threat. This demands comprehensive knowledge of all UAV systems for diagnostics and emergency protocols, together with advanced UAV-specific reasoning. The key challenge lies in achieving human pilot-level decision-making across operational scenarios.

\section{Conclusion}
\label{Sec. Conclusion}

This paper presented the Next-Generation LLM for UAV (\Name) system, which maps natural-language tasking to executable flight trajectories for short-, medium-, and long-range UAVs. The mapping is carried out by an end-to-end pipeline, in which the LLM interprets the pilot's instruction, route and path planning derive the waypoints, and the control platform generates the trajectory. We demonstrated the system on three case studies, namely short-range multi-UAV patrol, medium-range multi-POI delivery, and long-range multi-hop relocation, and evaluated it quantitatively. LLM-as-Parser proved substantially more accurate than the baselines it is compared against, and the system reached a mission success rate above 97\% on both medium- and long-range missions over 2000 airport pairs across the continental United States. In all three case studies, the system mapped a natural-language instruction to an executable trajectory in under three seconds, respecting user-defined constraints and the published airspace volumes.

Looking ahead, we outlined a five-level roadmap for LLM-powered UAV autonomy, progressing from the current LLM-as-Parser stage to the envisioned LLM-as-Autopilot with pilot-level decision making. This trajectory emphasizes the gradual integration of reasoning, planning, and control as models mature and as domain-specific datasets, regulatory codices, and communication protocols are incorporated. Key challenges remain, including the scarcity of UAV-specific datasets, the difficulty of embedding regulatory and safety knowledge into models, and the need for robust benchmarks and simulators for safety-critical evaluation. In conclusion, \Name serves both as a practical demonstration and as a forward-looking framework for LLM-enabled aerial systems, laying the foundation for UAVs that are safer, more adaptable, and more accessible to human operators.

\bibliographystyle{IEEEtran}
\small\bibliography{reference}

\vspace{-10mm}

\begin{IEEEbiography}
[{\includegraphics[width=1in,height=1.25in,clip,keepaspectratio]{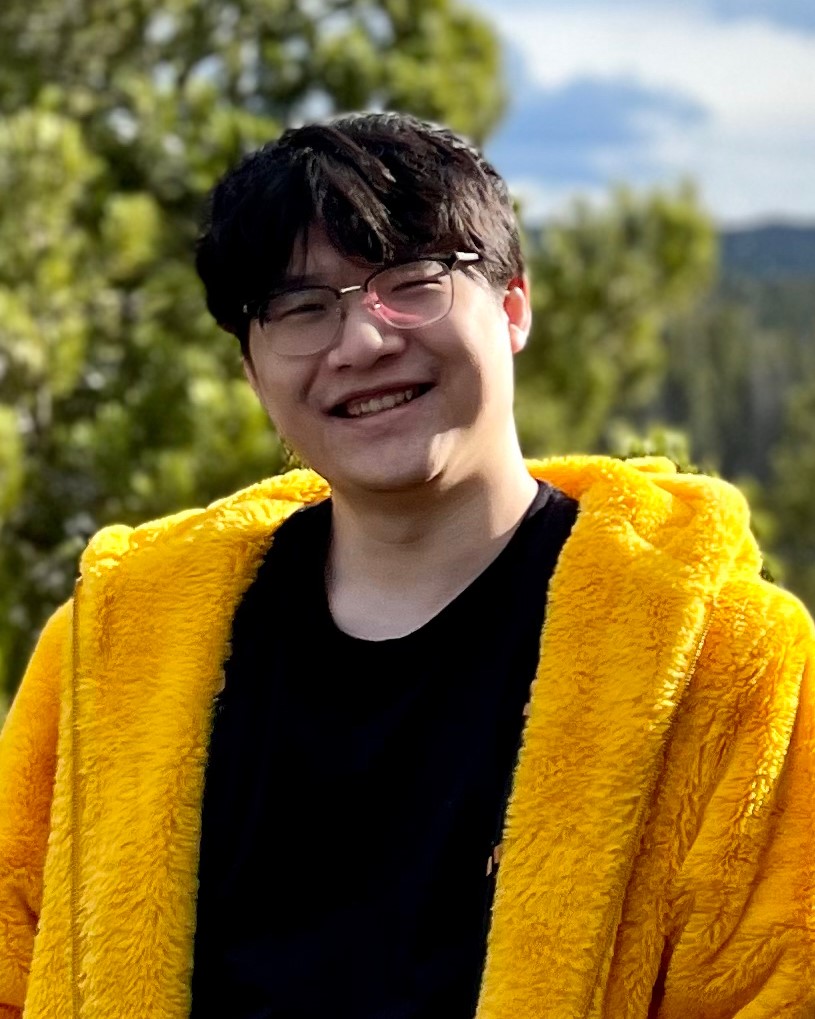}}]
{Liangqi Yuan}
received the B.E. degree in Photo-electronic Information Science and Engineering from Beijing Information Science and Technology University, Beijing, China, in 2020, and the M.S. degree in Electrical and Computer Engineering from Oakland University, Rochester, MI, USA, in 2022. He is currently pursuing the Ph.D. degree with the School of Electrical and Computer Engineering, Purdue University, West Lafayette, IN, USA. His research interests include multimodal learning, mobile computing, and machine learning.
\end{IEEEbiography}

\vspace{-10mm}

\begin{IEEEbiography}
[{\includegraphics[width=1in,height=1.25in,clip,keepaspectratio]{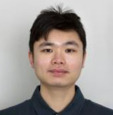}}]
{Chuhao Deng}
received the B.S. and M.S. degrees from Purdue University, West Lafayette, IN, USA, where he is currently pursuing the Ph.D. degree in aeronautics and astronautics. His research interests include air traffic management, human-machine interaction, and machine learning.
\end{IEEEbiography}

\vspace{-10mm}

\begin{IEEEbiography}
[{\includegraphics[width=1in,height=1.25in,clip,keepaspectratio]{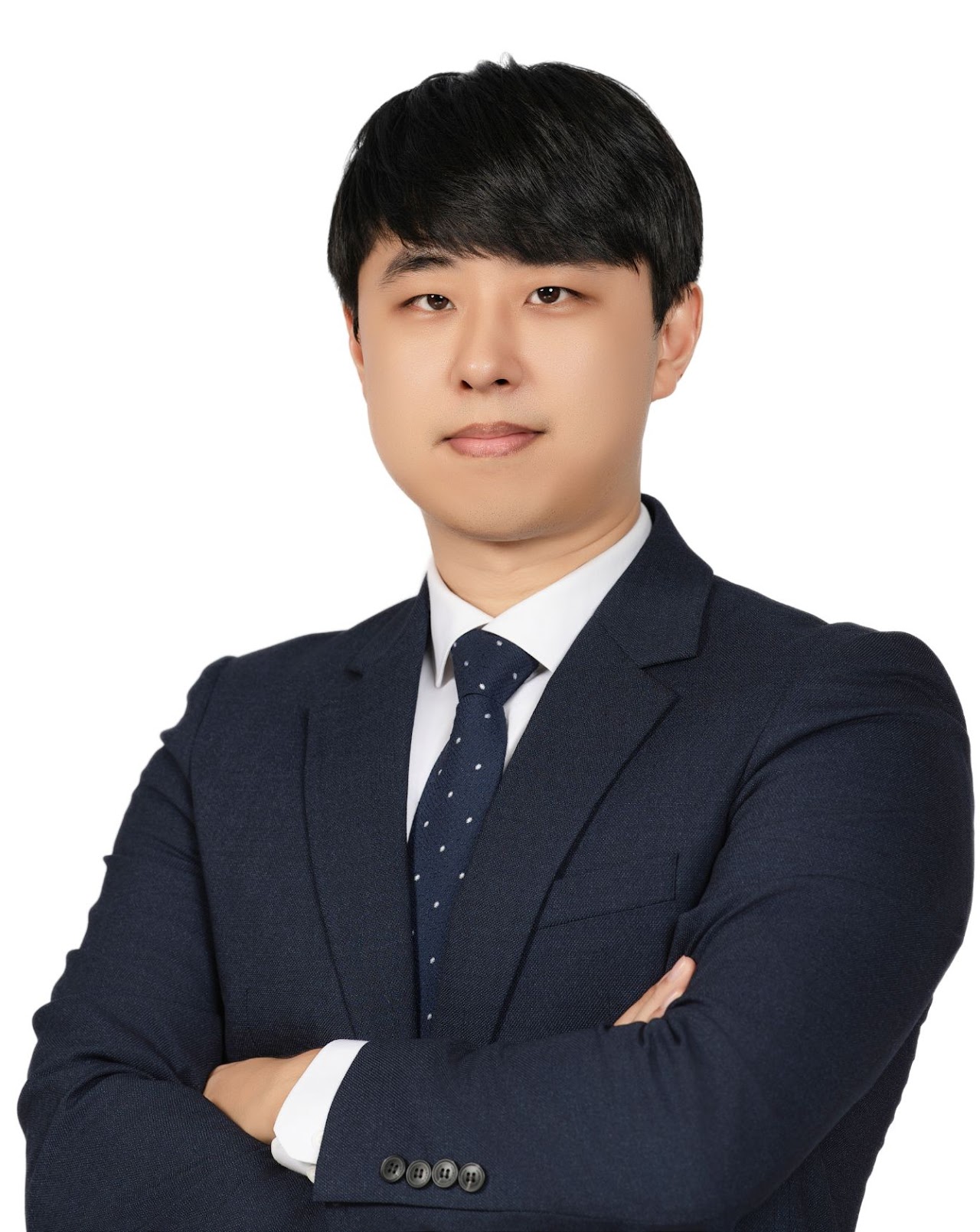}}] {Dong-Jun Han} 
received the B.S. degree in mathematics and electrical engineering and the M.S. and Ph.D. degrees in electrical engineering from KAIST, South Korea, in 2016, 2018, and 2022, respectively. He is currently an Assistant Professor with the Department of Computer Science and Engineering, Yonsei University, Seoul, South Korea. His research interests lie at the intersection of communications, networking, and machine learning, particularly distributed and federated learning and network optimization.
\end{IEEEbiography}

\vspace{-10mm}

\begin{IEEEbiography}
[{\includegraphics[width=1in,height=1.25in,clip,keepaspectratio]{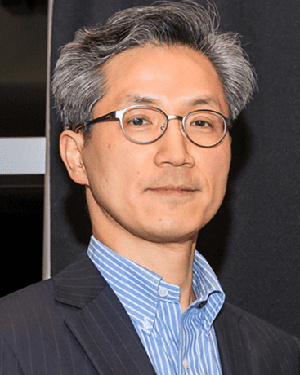}}]
{Inseok Hwang}
received the Ph.D. degree in aeronautics and astronautics from Stanford University, Stanford, CA, USA, in 2004. He is currently the Paul Stanley Professor with the School of Aeronautics and Astronautics, Purdue University, West Lafayette, IN, USA, where he leads the Flight Dynamics and Control/Hybrid Systems Laboratory. His research interests include high-assurance autonomy, which guarantees safety, security, and performance, for cyber-physical systems based on the hybrid systems approach, and its applications to safety-critical systems such as aircraft, spacecraft, unmanned aerial systems, air traffic management, and multiagent systems. Dr.~Hwang is an Associate Fellow of the American Institute of Aeronautics and Astronautics.
\end{IEEEbiography}

\vspace{-10mm}

\begin{IEEEbiography}
[{\includegraphics[width=1in,height=1.25in,clip,keepaspectratio]{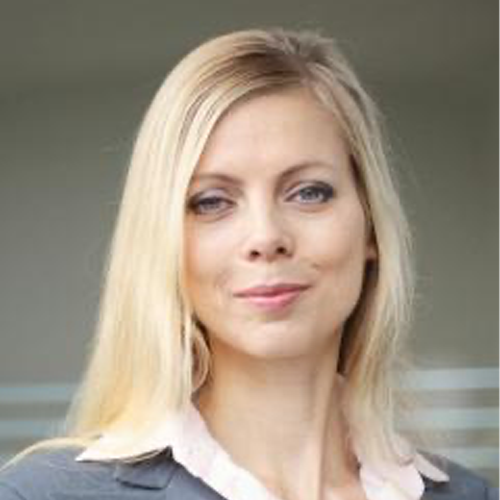}}]
{Sabine Brunswicker}
is a Full Professor at Purdue University, West Lafayette, IN, USA, and the Founder and Director of an interdisciplinary center of Artificial Intelligence for Digital, Autonomous and Augmented Aviation. Trained in systems engineering and decision sciences, she studies digital innovation, artificial intelligence, and autonomous systems. Her recent work addresses human-autonomy teaming between human operators and increasingly autonomous drones, and the role of emotional intelligence and empathy in conversations between humans and generative AI systems.
\end{IEEEbiography}

\vspace{-10mm}

\begin{IEEEbiography}
[{\includegraphics[width=1in,height=1.25in,clip,keepaspectratio]{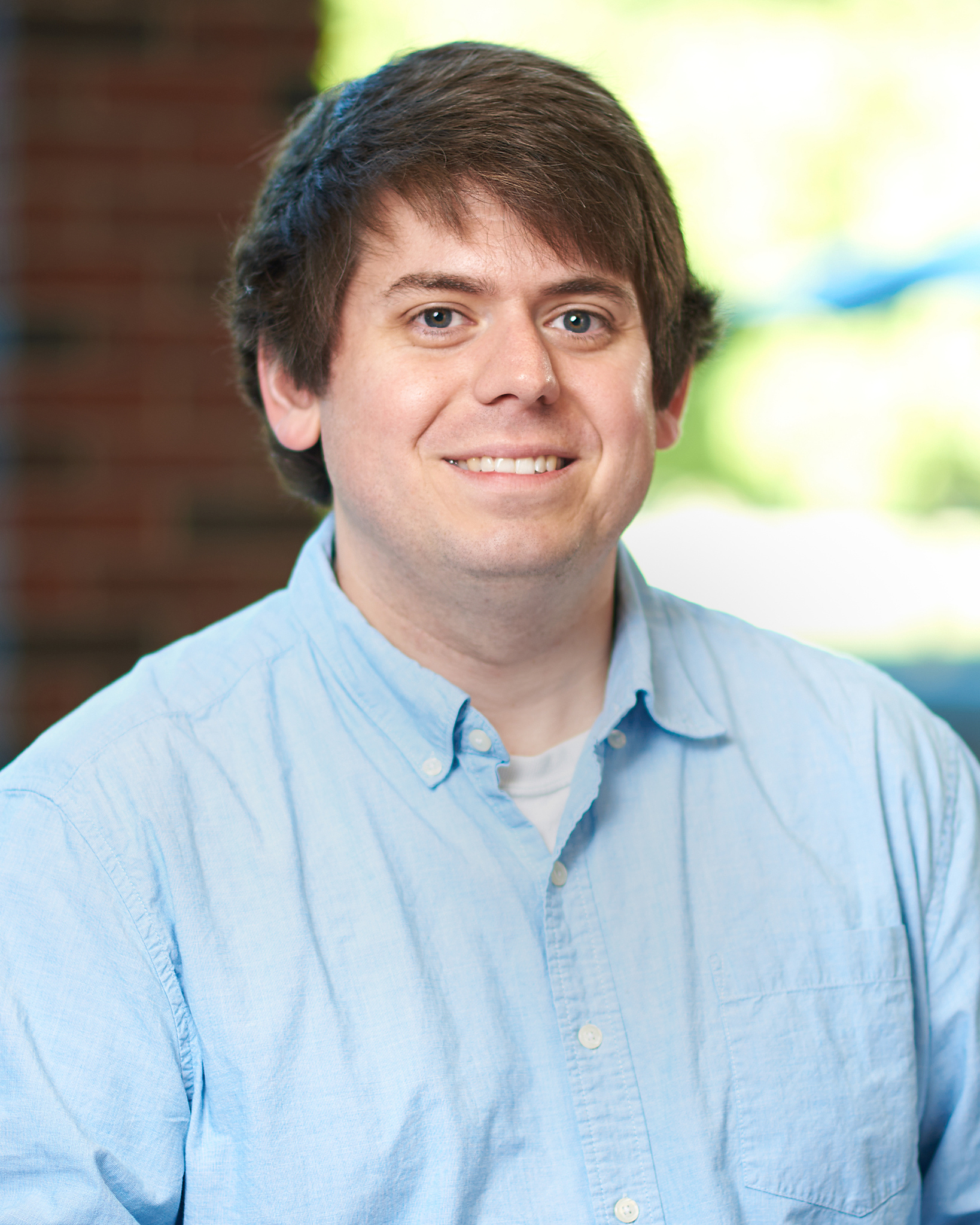}}]
{Christopher G. Brinton}
is the Elmore Associate Professor of ECE at Purdue University. He received his Ph.D. and M.Sc. in EE from Princeton University in 2016 and 2013, respectively. He is a recipient of four of the US top early career awards, from the National Science Foundation (CAREER), Office of Naval Research (YIP), Defense Advanced Research Projects Agency (YFA), and Air Force Office of Scientific Research (YIP). He is also a recipient of the Intel Rising Star Faculty Award and the Qualcomm Faculty Award.
\end{IEEEbiography}

\end{document}